%% file: iclr2027_conference.tex
\documentclass{article}
\usepackage{iclr2027_conference,times}
\input{math_commands.tex}

\usepackage{url}
\usepackage{amsmath,amssymb}
\usepackage{algorithm}
\usepackage{algorithmic}
\usepackage{graphicx}
\usepackage{booktabs}
\usepackage{colortbl}
\usepackage{wrapfig}
\usepackage{caption}
\usepackage{xcolor}
\usepackage[normalem]{ulem}
\usepackage{pifont}
\definecolor{citeblue}{rgb}{0.21,0.49,0.74}
\usepackage[pagebackref=false,breaklinks,colorlinks,citecolor=citeblue,bookmarks=false]{hyperref}
\definecolor{darkgreen}{RGB}{0,128,0}
\newcommand{\mycomment}[1]{\textcolor{darkgreen}{\textit{// #1}}}
\definecolor{tipgray}{RGB}{140,140,140}
\definecolor{posboxbg}{HTML}{f8fcf7}
\definecolor{posboxframe}{HTML}{8bc78a}
\definecolor{posboxtitle}{HTML}{5da05c}
\definecolor{negboxbg}{HTML}{faf5f6}
\definecolor{negboxframe}{HTML}{d9a0a8}
\definecolor{negboxtitle}{HTML}{c07080}
\newcommand{\tw}[2]{\colorbox{tipgray!#1}{\strut #2}}
\newcommand{\mk}[1]{\sout{#1}}

\usepackage[most,skins,theorems]{tcolorbox}
\usepackage{tcolorbox} 
\usepackage{listings}
\newtcolorbox{promptbox}[1][]{
  enhanced, breakable,
  colback=gray!1,      
  colframe=gray!60,    
  coltitle=black,      
  boxrule=2pt,
  arc=10pt,
  left=6pt, right=6pt, top=6pt, bottom=6pt,
  title={#1}, fonttitle=\bfseries,
  attach boxed title to top left={yshift*=-3mm},
  boxed title style={colback=gray!10}
}

\tcbuselibrary{breakable,skins}

\tcbset{
  aibox/.style={
    enhanced,
    breakable,                 %
    width=\linewidth,
    top=8pt,
    bottom=4pt,
    colback=inftythink-red!15,
    colframe=inftythink-red,
    colbacktitle=inftythink-red!90!black,
    attach boxed title to top left={yshift=-0.1in,xshift=0.15in},
    boxed title style={boxrule=0pt,colframe=white,},
    before upper={
      \parindent=0pt
      \raggedright
      \sloppy
      \emergencystretch=3em
    },
  }
}

\newtcolorbox{AIbox}[2][]{
  aibox,
  title=#2,
  notitle after break,          %
  #1
}

\newcommand{\methodname}{TTPO}
\newcommand{\fullname}{Test-Time Policy Optimization}
\title{TTPO: Test-Time Policy Optimization}

\author{\textbf{Aozhe Wang}$^{1,2,*}$,~
\textbf{Zhengxi Lu}$^{1,*}$,~
\textbf{Jianze Wang}$^{2}$,~
\textbf{Shangke Lv}$^{1}$,
\textbf{Ying Liu}$^{2}$,
\textbf{Weiming Lu}$^{1}$,
\\
\textbf{Jun Xiao}$^{1}$,~
\textbf{Yueting Zhuang}$^{1}$,~
\textbf{Hua Yang}$^{2}$,~
\textbf{Qianglong Chen}$^{2,\dagger}$,~
\textbf{Yongliang Shen}$^{1,\dagger}$
\\[3pt]
  $^1$Zhejiang University \qquad $^2$Alibaba Group\\
  \texttt{\small \{waz,zhengxilu,syl\}@zju.edu.cn \qquad qianglong.cql@alibaba-inc.com}
}

\iclrfinalcopy
\begin{document}
\include{figures/style}

\maketitle

\renewcommand{\thefootnote}{\fnsymbol{footnote}}
\footnotetext[1]{Equal contribution.}
\footnotetext[2]{Corresponding author.}
\renewcommand{\thefootnote}{\arabic{footnote}}
\setcounter{footnote}{0}

\vspace{-1.3mm}
\begin{abstract}
Recent prominent post-training methods, such as Reinforcement Learning (RL) and On-Policy Self-Distillation (OPSD), have driven rapid progress in mathematical reasoning for large language models, yet their reliance on ground-truth labels precludes test-time training (TTT). Replacing ground truth with majority-vote pseudo-labels is a natural alternative, yet it is fragile: an incorrect vote corrupts the teacher and misleads every token. We observe that this failure mode is asymmetric: rollouts that disagree with the pseudo-label are typically wrong regardless of whether the vote itself is correct. Building on this observation, we propose \textbf{\fullname{} (\methodname{})}, an asymmetric objective that distills agreeing rollouts via OPSD and penalizes disagreeing rollouts with Grouped RL. Token-level selection further refines both branches: distillation down-weights already-converged positions, while RL penalizes only confident errors. Both updates remain well-grounded even under frequent pseudo-label errors, and majority-vote routing yields tighter self-supervision as the model improves. Without any labels, \methodname{} matches label-supervised OPSD on five competition-level benchmarks, raises Qwen3-1.7B from 38.0\% to 45.2\% in TTT, yields +25.2\% to +36.4\% without thinking, and shows strong cross-task generalization. Our code is available at \url{https://github.com/ZJU-REAL/TTPO}.

\end{abstract}

\vspace{-1mm}

\begin{figure}[h]
\centering
\includegraphics[width=\linewidth]{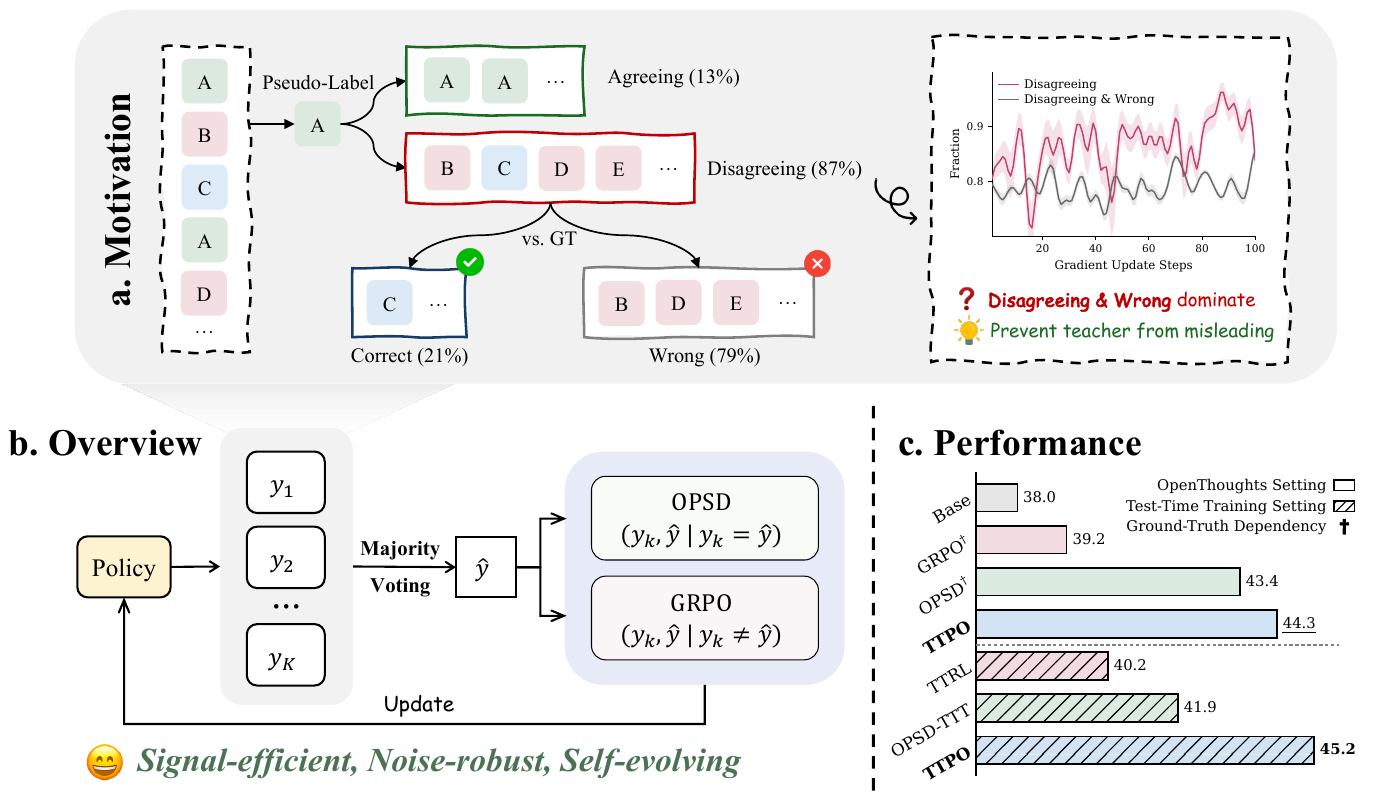}
\caption{\textbf{(a) Motivation:}~When pseudo-labels are wrong, most disagreeing rollouts are genuinely wrong too, during Qwen3-1.7B TTT on AIME 2026. \textbf{(b) Overview of \methodname{}}. \textbf{(c) Performance:}~Average accuracy of Qwen3-1.7B across AIME 2026, HMMT 2026, and BRUMO 2025.}
\vspace{-0.5mm}
\label{fig:intro}
\end{figure}

\section{Introduction}
\label{sec:intro}

Large language models (LLMs) have achieved remarkable mathematical reasoning through extended chain-of-thought generation~\citep{team2026kimi,xu2026deepseek,zeng2026glm,singh2025openai,team2026qwen3}, largely powered by post-training with reinforcement learning from verifiable rewards (RLVR)~\citep{shao2024deepseekmath, yu2026dapo, guo2025deepseek}. Yet such outcome rewards are inherently coarse, broadcasting a single sequence-level scalar uniformly across all tokens and leaving the reasoning at each step unsupervised.

A complementary line supplies dense, token-level supervision: on-policy self-distillation (OPSD)~\citep{zhao2026self} conditions the same policy on the ground-truth answer to form a teacher that re-scores the student's own rollouts token by token~\citep{ye2026policy, yang2026learning}. Recent work increasingly combines the two signals, through auxiliary distillation losses, credit redistribution, or routing between them~\citep{yang2026self,liu2026sdpg,lu2026sdar,han2026rstg,li2026unifying}. All these methods, however, assume ground-truth answers: the reward needs them for verification, and the teacher needs them as privileged context. In \emph{test-time training} (TTT)~\citep{sun2020test}, where a model improves on the very problems it must solve and labels never arrive, none of these methods applies.

Without labels, supervision must come from the model itself: sample a group of rollouts for each problem, and take the majority answer as a pseudo-label. TTRL~\citep{zuo2026ttrl} uses this pseudo-label as a reward and improves reasoning without any labels, but the reward is still one scalar per trajectory, and when the majority is wrong, training reinforces the error~\citep{lin2026extinction}. The natural next step is to let the pseudo-label replace the ground-truth answer in OPSD, which has been tried in two forms: distilling all rollouts toward the pseudo-label-conditioned teacher~\citep{gkountouras2026canon}, or distilling only the rollouts that disagree with the pseudo-label~\citep{li2026policy}. But dense supervision magnifies label errors: \textbf{a corrupted reward misleads once per trajectory; a corrupted teacher misleads at every token}.

These errors are the common case: on competition-level problems, the pseudo-label is wrong for ${\sim}$85\% of prompts (Figure~\ref{fig:intro}, a). Learning from such a label seems infeasible. \textbf{However, even when the pseudo-label is wrong, ${\sim}$79\% of the rollouts that disagree with it are wrong too.} A penalty on a disagreeing rollout is therefore usually correct whether or not the pseudo-label is, because it uses the disagreement alone, never the pseudo-label's answer. Distillation toward the pseudo-label has no such tolerance: a wrong answer enters the teacher and misleads every token. The same asymmetry underlies negative learning from noisy labels, where stating what a sample is not remains reliable even when the label is wrong~\citep{kim2019nlnl}.

We propose \textbf{\fullname{}} (\methodname{}), which applies each signal where it is reliable: GRPO penalties on the rollouts that disagree with the pseudo-label, and OPSD distillation on the rollouts that agree with it. The distillation branch tolerates wrong pseudo-labels for a different reason: the teacher is conditioned on the answer that the agreeing rollouts themselves produced, so even when that answer is wrong, the update distills the model's thinking mode into its non-thinking mode rather than toward an arbitrary error. Finally, token-level selection sharpens both branches, weighting distillation toward positions the student has not yet mastered and masking penalties to the confident errors that caused the failure. By combining both signals, our asymmetric design not only makes more effective use of both positive and negative rollouts while remaining robust to pseudo-label noise, but is also naturally calibrated to the model's current capability, enabling a virtuous cycle of self-evolution that ground-truth routing cannot sustain (Figure~\ref{fig:upper_bound},~\ref{fig:avg_maj}).

Trained without any labels, \methodname{} matches or exceeds label-supervised OPSD across Qwen3-1.7B/4B/8B on five competition-level benchmarks, and in the pure TTT setting raises the 1.7B base model from 38.0\% to 45.2\% average accuracy, ahead of both TTRL and self-distillation baselines.
With thinking mode disabled, the gains reach +25.2\% to +36.4\% across scales, several times the gain of label-supervised OPSD. We further validate that training on any one benchmark improves the other two, indicating generalizable reasoning rather than problem-specific overfitting (Figure~\ref{fig:generalization}). Our contributions are:
\begin{enumerate}
    \item We show that majority-vote pseudo-labels remain useful despite frequent errors: though wrong on ${\sim}$85\% of competition-level prompts, ${\sim}$79\% of disagreeing rollouts are wrong too, so penalizing disagreement stays correct while distillation does not.

    \item We propose \methodname{}, which applies each signal where it stays correct: agreeing rollouts are distilled toward an answer-conditioned teacher, disagreeing rollouts receive GRPO penalties, with token-level selection in both branches.

    \item Trained without any labels, \methodname{} matches or exceeds label-supervised OPSD on five competition-level benchmarks, raises Qwen3-1.7B from 38.0\% to 45.2\% in TTT, and further demonstrates strong cross-task generalization.

\end{enumerate}

\section{Related Work}
\label{sec:related}

\subsection{Test-Time Training for Reasoning}

Test-time training (TTT) adapts models to unlabeled test data at inference time~\citep{sun2020test,li2026policy,du2025ttrl-gui}. TTRL~\citep{zuo2026ttrl} extends TTT to LLM reasoning by sampling multiple trajectories per problem, deriving pseudo-rewards via majority voting, and training with GRPO~\citep{shao2024deepseekmath}. Follow-up work addresses TTRL's sensitivity to consensus quality: Hi-TTRL~\citep{xu2026hi} introduces hierarchical reward shaping with hints, while SCRL~\citep{yan2026if} applies selective pseudo-labeling to filter unreliable majorities. However, these methods remain purely RL-based, propagating a single sequence-level reward uniformly across all tokens.

\subsection{On-Policy Self-Distillation}

On-policy distillation trains a policy on its own rollouts under a teacher~\citep{agarwal2024gkd,gu2026minillm,wen2023fdivergence}. Recent self-distillation variants remove the need for a separate teacher by conditioning the same model on privileged information available only during training~\citep{zhao2026self,he2026sdzero,lu2026skill0}. Several studies further incorporate the resulting teacher--student log-probability gap into RLVR, either as advantage scaling~\citep{yang2026self}, a detached auxiliary objective~\citep{lu2026sdar}, a routing mechanism~\citep{han2026rstg,li2026unifying}, or reward-densifying local supervision~\citep{xu2026tip,ye2026policy,he2026sdzero}. We follow this idea and introduce an asymmetric objective that decouples the treatment of positive and negative samples to tolerate pseudo-label errors.

\subsection{Token-Level Weighting and Masking}

Recent work recognizes that not all tokens merit equal gradients during training~\citep{xiao2026finding}. In distillation, TIP~\citep{xu2026tip} shows that training on fewer than 10\% of tokens, selected by student entropy and teacher--student divergence, nearly matches full-token performance. In RL, STAPO~\citep{liu2026stapo} masks spurious low-probability, low-entropy tokens in positive samples that receive disproportionate reward gradients, while \citet{wu2026rethinking} address the erroneous penalization of locally correct tokens within failed trajectories through reward recalibration. \methodname{} applies token-level selection to both branches of its objective: down-weighting converged positions in the distillation branch and selectively penalizing only confident errors in the RL branch.

\section{Method}
\label{sec:method}

\begin{figure}[t]
\centering
\includegraphics[width=\textwidth]{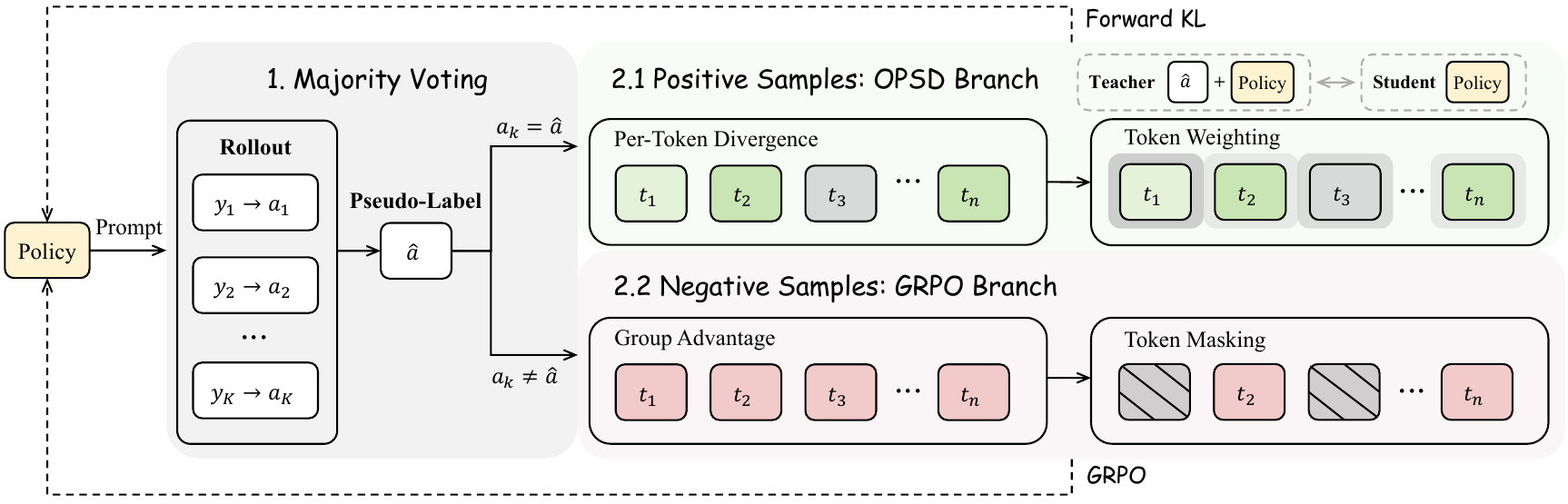}
\caption{\textbf{Overview of \methodname{}.} (1) Majority Voting: $K$ sampled trajectories are partitioned into positive ($a_k = \hat{a}$) and negative ($a_k \neq \hat{a}$) sets. (2.1) OPSD Branch: positive samples are supervised via per-token forward KL with token weighting. (2.2) GRPO Branch: negative samples are penalized via group advantages with token masking on anomalous positions.}
\label{fig:method}
\end{figure}

\subsection{Preliminaries and Problem Setup}
\label{sec:method:setup}

Let $\pi_\theta$ denote the language model and $\{x_i\}_{i=1}^N$ be a set of test-time problems without ground-truth labels. For each problem $x$, we sample $K$ trajectories $\{y_1, \ldots, y_K\} \sim \pi_\theta(\cdot \mid x)$ and extract final answers $a_k = \operatorname{Extract}(y_k)$.

\paragraph{Majority-vote pseudo-labeling.} We cluster answers by mathematical equivalence, select the largest cluster as the pseudo-label $\hat{a}$ with consensus count $c = |\{k : a_k \equiv \hat{a}\}|$, and partition trajectories into positive samples $\mathcal{P} = \{k : a_k \equiv \hat{a}\}$ that agree with pseudo-label and negative samples $\mathcal{N} = \{k : a_k \not\equiv \hat{a}\}$ that disagree.

\paragraph{Answer-conditioned teacher.} Following OPSD~\citep{zhao2026self}, we construct a teacher by conditioning the \emph{same} model on the pseudo-label $\hat{a}$ as privileged information. Given shared completion tokens $y_k$, the teacher and student differ only in their prompt prefixes:
\begin{align}
    q_t^{(\hat{a})} &= \pi_\theta(\cdot \mid [x; \hat{a}]_{\text{teacher}}, y_{<t}), \quad \text{(no grad)} \\
    p_t &= \pi_\theta(\cdot \mid x_{\text{student}}, y_{<t}). \quad \text{(with grad)}
\end{align}

\subsection{Motivation: Why Asymmetric?}
\label{sec:method:motivation}

In the TTT setting, the training set \emph{is} the test set — consisting of competition-level problems that are inherently difficult for the model. As a result, majority-vote pseudo-labels are frequently wrong: on AIME 2026 with Qwen3-1.7B, we observe that pseudo-labels are wrong for ${\sim}$85\% of prompts on average (Figure~\ref{fig:intro}, a). Naively applying self-distillation to all trajectories using the corrupted teacher would propagate errors to every sample.

However, we observe a key structural property: even when $\hat{a} \neq a^*$, ${\sim}$79\% of negative samples produce answers that are neither $\hat{a}$ nor $a^*$ — penalizing them is correct regardless of pseudo-label quality. This motivates an \textbf{asymmetric} design that minimizes the blast radius of pseudo-label errors:
\begin{itemize}
    \item \textbf{Pure FKL on all samples}: error propagates to all $K$ trajectories (the teacher distribution is entirely conditioned on wrong $\hat{a}$).
    \item \textbf{Asymmetric (FKL for $\mathcal{P}$, GRPO for $\mathcal{N}$)}: only the small $|\mathcal{P}|$ set is corrupted. GRPO on $\mathcal{N}$ requires only ``not in the majority cluster'' — independent of $\hat{a}$'s content — and is correct for the vast majority of negative samples.
\end{itemize}

\begin{algorithm}[t]
\caption{\methodname{} Training}
\label{alg:ttpo}
\begin{algorithmic}[1]
\REQUIRE Policy $\pi_\theta$; problems $\{x_i\}$; rollouts $K$; RL weight $\lambda$
\FOR{each training iteration}
    \STATE \mycomment{Step 1: Majority-vote pseudo-labeling}
    \FOR{each problem $x$}
        \STATE $y^{(1)}, \ldots, y^{(K)} \sim \pi_\theta(\cdot \mid x)$; \quad $a^{(k)} \leftarrow \operatorname{Extract}(y^{(k)})$
        \STATE $\hat{a} \leftarrow \operatorname{plurality}(\{a^{(k)}\})$; \quad $\mathcal{P} \leftarrow \{k : a^{(k)} = \hat{a}\}$; \quad $\mathcal{N} \leftarrow \{k : a^{(k)} \neq \hat{a}\}$
    \ENDFOR
    \STATE \mycomment{Step 2: Teacher \& student forward}
    \STATE $q_t \leftarrow \pi_\theta(\cdot \mid [x;\hat{a}], y_{<t})$ \textcolor{gray}{\small(no grad)}; \quad $p_t \leftarrow \pi_\theta(\cdot \mid x, y_{<t})$
    \STATE \mycomment{Step 3: OPSD on $\mathcal{P}$}
    \STATE $w(t) \leftarrow \hat{H}(t) + \hat{\Delta}(t) - \hat{H}(t)\cdot\hat{\Delta}(t)$
    \STATE $\mathcal{L}_{\text{OPSD}} \leftarrow \frac{1}{|\mathcal{P}|}\sum_{k \in \mathcal{P}} \frac{1}{T_k}\sum_{t} w(t) \cdot \mathrm{KL}(q_t \| p_t)$
    \STATE \mycomment{Step 4: GRPO on $\mathcal{N}$}
    \STATE $s(t) \leftarrow -\log p_t(y^{(t)}_k) \cdot (1 - \hat{H}(t))$; \quad $m(t) \leftarrow \mathbf{1}[s(t) \geq \mathrm{median}(\{s\})]$
    \STATE $\mathcal{L}_{\text{GRPO}} \leftarrow -\frac{1}{|\mathcal{N}|}\sum_{k \in \mathcal{N}} \frac{A_k}{T_k}\sum_{t} m(t) \cdot \log \pi_\theta(y^{(t)}_k \mid x, y^{(<t)}_k)$
    \STATE \mycomment{Step 5: Update}
    \STATE $\theta \leftarrow \theta - \eta\,\nabla_\theta\!\left(\mathcal{L}_{\text{OPSD}} + \lambda\,\mathcal{L}_{\text{GRPO}}\right)$
\ENDFOR
\end{algorithmic}
\end{algorithm}

\subsection{Positive Samples: OPSD Branch}
\label{sec:method:positive}

For each positive sample $y_k \in \mathcal{P}$, we apply OPSD's forward KL with per-token weighting:
\begin{equation}
    \mathcal{L}_{\text{OPSD}}(k) = \frac{1}{T_k} \sum_{t=1}^{T_k} w(t) \cdot \text{KL}\left(q_t^{(\hat{a})} \,\|\, p_t\right),
\end{equation}
where $T_k$ is the number of valid response tokens and $w(t)$ is the token weight described below.

\paragraph{Token weighting.} Not all tokens offer equal learning value. Inspired by the token importance analysis of~\citet{xu2026tip}, we design a weighting scheme that down-weights positions where the student has already converged. We measure two complementary signals — student entropy $H(t)$ and teacher-student divergence $\Delta(t) = \text{KL}(q_t \| p_t)$ — normalize each to $[0,1]$ via per-sample min-max normalization, and combine them with a Soft-OR:
\begin{equation}
    w(t) = \hat{H}(t) + \hat{\Delta}(t) - \hat{H}(t) \cdot \hat{\Delta}(t).
    \label{eq:tip}
\end{equation}
This assigns high weight when \emph{at least one} signal indicates learning value (the student is uncertain, or confidently wrong), and approaches zero only when both signals are low — the student is already confident and aligned with the teacher.

\subsection{Negative Samples: GRPO Branch}
\label{sec:method:negative}

For negative samples $y_k \in \mathcal{N}$, we apply GRPO~\citep{shao2024deepseekmath}. Each trajectory receives a binary reward based on majority-vote classification: $r_k = 1$ if its answer matches the pseudo-label, $r_k = 0$ otherwise. Advantages are computed group-relatively over all $K$ rollouts per problem, so that $A_k < 0$ for negative samples, yielding a penalty:
\begin{equation}
    \mathcal{L}_{\text{GRPO}}(k) = -\frac{A_k}{T_k} \sum_{t=1}^{T_k} m(t) \cdot \log \pi_\theta(y_k^{(t)} \mid x, y_k^{(<t)}).
\end{equation}

In standard GRPO, positive-advantage gradients counterbalance erroneous penalties on locally correct tokens within failed trajectories. In our asymmetric framework, GRPO acts \emph{exclusively} on negative samples, removing this counterbalance. This \emph{False Penalties on Negative Samples} problem~\citep{wu2026rethinking} necessitates active reduction of collateral damage.

\paragraph{Token masking.} Drawing on the insight from~\citet{liu2026stapo} that low-probability, low-entropy tokens represent anomalous model behavior, we design a masking scheme to identify tokens most responsible for errors. We score each token by:
\begin{equation}
    s(t) = -\log \pi_\theta(y_k^{(t)} \mid x, y_k^{(<t)}) \cdot (1 - \hat{H}(t)),
    \label{eq:anomaly_score}
\end{equation}
where the raw negative log-probability serves as the dominant ranking factor and $(1 - \hat{H}(t))$ is the normalized certainty. The key design choice is using unnormalized $-\log p$ to anchor the ranking: this ensures locally correct tokens (typically high-probability) are naturally excluded, while genuinely anomalous outputs — where the model confidently produced unlikely content — are prioritized. We construct a binary mask by selecting the top-50\% of tokens by score:
\begin{equation}
    m(t) = \mathbf{1}\left[s(t) \geq \text{median}(\{s(t')\}_{t'=1}^{T_k})\right].
\end{equation}

\subsection{Unified Objective}
\label{sec:method:unified}

The final \methodname{} objective combines both branches, and balances their weight by $\lambda$:
\begin{equation}
    \mathcal{L}_{\text{TTPO}} = \frac{1}{|\mathcal{B}|} \left( \sum_{k \in \mathcal{P}} \mathcal{L}_{\text{OPSD}}(k) + \lambda \sum_{k \in \mathcal{N}} \mathcal{L}_{\text{GRPO}}(k) \right),
    \label{eq:ttpo_loss}
\end{equation}
The complete training procedure is summarized in Algorithm~\ref{alg:ttpo}.

\section{Experiments}
\label{sec:exp}

\subsection{Experimental Setup}
\label{sec:exp:setup}

\paragraph{Implementation.} We evaluate on Qwen3-1.7B, Qwen3-4B, and Qwen3-8B~\citep{yang2025qwen3}, all fine-tuned with LoRA ($r{=}64$, $\alpha{=}128$) on all linear layers. We consider two settings: (1) \textbf{OpenThoughts setting}, where models are trained on labeled data but \methodname{} does not use the labels — they serve only for comparison with label-dependent baselines; and (2) \textbf{TTT setting}, where models are trained directly on the test set without any annotations. The shared training configuration follows OPSD, and full training configurations are provided in Appendix~\ref{app:impl}.

\paragraph{Baselines.} We compare against: (1) \textbf{OPSD}~\citep{zhao2026self}, on-policy self-distillation with ground-truth labels; (2) \textbf{GRPO}~\citep{shao2024deepseekmath}, RL with ground-truth rewards; (3) \textbf{TTRL}~\citep{zuo2026ttrl}, label-free RL via majority-vote rewards; and (4) \textbf{OPSD-TTT}, self-distillation using the model's temperature-0 output under thinking mode as privileged information.

\paragraph{Evaluation.} We evaluate on five competition-level math benchmarks: AIME 2025, AIME 2026, HMMT 2025, HMMT 2026, and BRUMO 2025~\citep{dekoninck2026matharena}. To ensure train-inference consistency, evaluation is performed with thinking mode enabled (non-thinking evaluation in Appendix~\ref{app:nonthinking}). All results are reported as Avg@12 with temperature 1.0. For OPSD and \methodname{}, we train for 100 steps and report the peak performance across checkpoints saved every 25 steps. For GRPO and TTRL, we train for 500 steps and report the peak across all checkpoints.

\paragraph{TTPO-specific hyperparameters.} We sample $K{=}64$ trajectories per problem to ensure reliable majority voting on hard problems with low pass rates, with a maximum generation length of 16,000 tokens to avoid truncation that prevents answer extraction. From the $K$ rollouts, $K_{\text{train}}{=}8$ are selected (50\% positive, 50\% negative) for the gradient update, and the RL weight $\lambda{=}0.1$ balances gradient magnitudes between the two branches (ablated in Appendix~\ref{app:ablation}).

\subsection{Main Results}
\label{sec:exp:main}

\begin{table}[t]
\centering
\small
\caption{Results on OpenThoughts training data. OPSD and GRPO use ground-truth labels ($\dagger$); \methodname{} uses only majority-vote pseudo-labels.}
\label{tab:main_openthoughts}
\begin{tabular}{p{2.4cm}|cccccc}
\toprule
\textbf{Method} & \textbf{AIME25} & \textbf{HMMT25} & \textbf{AIME26} & \textbf{HMMT26} & \textbf{BRUMO25} & \textbf{Average} \\
\midrule
\multicolumn{7}{@{}l}{\textit{Qwen3-1.7B}} \\
\hspace{0.8em}Base & 36.9 & 21.9 & 37.8 & 28.8 & 47.5 & 34.6 \\
\hspace{0.8em}+GRPO$^\dagger$ & 37.3 & 23.6 & 40.3 & 29.3 & 48.1 & 35.7 \\
\hspace{0.8em}+OPSD$^\dagger$ & \underline{40.3} & \textbf{28.1} & \underline{46.4} & \underline{31.4} & \underline{52.5} & \underline{39.7} \\
\rowcolor[HTML]{E4F0F9} \hspace{0.8em}+\methodname{} & \textbf{41.7} & \underline{26.1} & \textbf{46.5} & \textbf{31.6} & \textbf{54.7} & \textbf{40.1} \\
\midrule
\multicolumn{7}{@{}l}{\textit{Qwen3-4B}} \\
\hspace{0.8em}Base & 66.1 & 41.9 & 65.8 & 42.4 & 64.0 & 56.0 \\
\hspace{0.8em}+GRPO$^\dagger$ & 66.7 & \textbf{45.0} & \underline{66.6} & 43.2 & \underline{66.1} & 57.5 \\
\hspace{0.8em}+OPSD$^\dagger$ & \underline{68.3} & \underline{44.2} & \textbf{68.1} & \underline{44.4} & \textbf{67.2} & \underline{58.4} \\
\rowcolor[HTML]{E4F0F9} \hspace{0.8em}+\methodname{} & \textbf{69.4} & 43.6 & \textbf{68.1} & \textbf{44.7} & \textbf{67.2} & \textbf{58.6} \\
\midrule
\multicolumn{7}{@{}l}{\textit{Qwen3-8B}} \\
\hspace{0.8em}Base & 66.7 & 44.2 & 67.5 & 45.5 & 69.2 & 58.6 \\
\hspace{0.8em}+GRPO$^\dagger$ & 70.3 & \textbf{46.7} & 69.2 & \textbf{48.0} & \underline{71.9} & 61.2 \\
\hspace{0.8em}+OPSD$^\dagger$ & \underline{70.8} & \underline{46.4} & \underline{72.5} & 47.2 & 71.4 & \underline{61.7} \\
\rowcolor[HTML]{E4F0F9} \hspace{0.8em}+\methodname{} & \textbf{71.4} & 46.1 & \textbf{74.2} & \textbf{48.0} & \textbf{73.1} & \textbf{62.6} \\
\bottomrule
\end{tabular}
\end{table}

\begin{table}[t]
\centering
\small
\caption{Results on test-time training data. No method uses ground-truth labels.}
\label{tab:main_ttt}
\begin{tabular}{p{2.4cm}|cccc}
\toprule
\textbf{Method} & \textbf{AIME26} & \textbf{HMMT26} & \textbf{BRUMO25} & \textbf{Average} \\
\midrule
\multicolumn{5}{@{}l}{\textit{Qwen3-1.7B}} \\
\hspace{0.8em}Base & 37.8 & 28.8 & 47.5 & 38.0 \\
\hspace{0.8em}+TTRL & 39.2 & \underline{30.6} & \underline{50.9} & 40.2 \\
\hspace{0.8em}+OPSD-TTT & \underline{44.7} & 30.3 & 50.8 & \underline{41.9} \\
\rowcolor[HTML]{E4F0F9} \hspace{0.8em}+\methodname{} & \textbf{48.9} & \textbf{33.6} & \textbf{53.1} & \textbf{45.2} \\
\midrule
\multicolumn{5}{@{}l}{\textit{Qwen3-4B}} \\
\hspace{0.8em}Base & 65.8 & 42.4 & 64.0 & 57.4 \\
\hspace{0.8em}+TTRL & 66.4 & 43.2 & \underline{66.7} & 58.8 \\
\hspace{0.8em}+OPSD-TTT & \underline{67.8} & \underline{43.4} & \textbf{66.9} & \underline{59.4} \\
\rowcolor[HTML]{E4F0F9} \hspace{0.8em}+\methodname{} & \textbf{70.8} & \textbf{45.7} & \textbf{66.9} & \textbf{61.1} \\
\midrule
\multicolumn{5}{@{}l}{\textit{Qwen3-8B}} \\
\hspace{0.8em}Base & 67.5 & 45.5 & 69.2 & 60.7 \\
\hspace{0.8em}+TTRL & 70.8 & \underline{48.0} & 70.1 & 63.0 \\
\hspace{0.8em}+OPSD-TTT & \underline{71.7} & 47.2 & \underline{72.2} & \underline{63.7} \\
\rowcolor[HTML]{E4F0F9} \hspace{0.8em}+\methodname{} & \textbf{73.9} & \textbf{48.5} & \textbf{73.6} & \textbf{65.3} \\
\bottomrule
\end{tabular}
\end{table}

\paragraph{Labeled training data.}
Table~\ref{tab:main_openthoughts} compares methods trained on OpenThoughts, where OPSD and GRPO use ground-truth labels while \methodname{} relies solely on majority-vote pseudo-labels. \methodname{} exceeds the label-dependent OPSD across all three model scales (40.1 vs.\ 39.7 on 1.7B, 58.6 vs.\ 58.4 on 4B, 62.6 vs.\ 61.7 on 8B in average), despite without ground-truth supervision. This demonstrates that majority-vote pseudo-labels, when combined with our asymmetric objective, can substitute for ground-truth annotations without sacrificing performance. The improvements are consistent across scales, and notably, \methodname{} on Qwen3-4B (58.6 avg) already matches the Qwen3-8B base model (58.6 avg), suggesting that our training recipe effectively amplifies a smaller model's reasoning capacity to the level of a $2\times$ larger untrained model.

\paragraph{Label-free test-time training.}
Table~\ref{tab:main_ttt} evaluates the purely label-free TTT setting where models train directly on the test problems. \methodname{} consistently and substantially outperforms both TTRL and OPSD-TTT across all model scales. On Qwen3-1.7B, \methodname{} achieves 45.2 average --- +3.3 over OPSD-TTT and +5.4 over TTRL, representing a 7.2-point absolute gain over the base model. The gap over TTRL demonstrates the value of dense distributional guidance: while TTRL provides only binary reward signals, \methodname{} additionally leverages the answer-conditioned teacher to transfer token-level knowledge on correct trajectories. The gap over OPSD-TTT --- which uses deterministic (greedy decoding with thinking mode enabled) answers as privileged information rather than majority-vote pseudo-labels --- shows that even with a reasonable self-distillation baseline, our asymmetric design extracts substantially more signal by additionally exploiting negative samples through selective RL penalties. Cross-scale comparison further highlights the efficiency: \methodname{} on Qwen3-4B (61.1 avg) already surpasses Qwen3-8B base (60.7 avg), demonstrating that label-free test-time training with \methodname{} can close the gap between model sizes.

\subsection{Ablation Studies}
\label{sec:exp:ablation}
\paragraph{Token-level selection.}

\begin{wraptable}{r}{0.56\textwidth}
\vspace{-13pt}
\centering
\small
\caption{Token-level selection ablation results on Qwen3-1.7B OpenThoughts setting.}
\label{tab:ablation_weight}
\begin{tabular}{l|ccc}
\toprule
\textbf{Method} & \textbf{AIME26} & \textbf{HMMT26} & \textbf{BRUMO25} \\
\midrule
\rowcolor[HTML]{E4F0F9} \methodname{} & \textbf{46.5} & \textbf{31.6} & \textbf{54.7} \\
\hspace{0.3em}w/o pos.\ weight & 43.3 & 30.6 & 52.8 \\
\hspace{0.3em}w/o neg.\ mask & 45.4 & 29.5 & 50.0 \\
\bottomrule
\end{tabular}
\vspace{-8pt}
\end{wraptable}

Table~\ref{tab:ablation_weight} isolates the contribution of each token-level selection mechanism. Both components improve over uniform updates, but their effects are complementary and target different failure modes: removing positive-sample weighting (w/o pos.\ weight) uniformly distills all tokens including low-value (low-entropy, low-divergence) positions where the student has already converged, diluting the gradient signal from genuinely informative tokens; removing negative-sample masking (w/o neg.\ mask) penalizes all tokens indiscriminately --- not only causing collateral damage to locally correct reasoning steps that cannot be offset without positive-advantage updates, but also allowing anomalous (low-probability, low-entropy) tokens to dominate gradient updates, injecting substantial noise into optimization. The full method benefits from both --- focusing distillation where it matters and penalizing only where errors originate.

\paragraph{Update strategy.}

\begin{figure}[t]
\centering
\includegraphics[width=\textwidth]{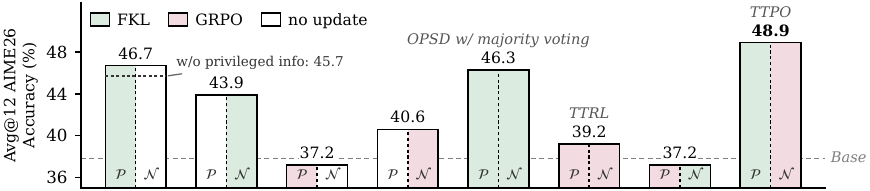}
\caption{\textbf{Ablations over update strategies on Qwen3-1.7B AIME26 TTT setting.} $\mathcal{P}$ and $\mathcal{N}$ denote the loss applied to positive and negative samples, respectively. Dotted lines indicate the corresponding variants with an unconditioned teacher.}
\label{fig:ablation_update}
\end{figure}

Figure~\ref{fig:ablation_update} compares update strategy combinations. The full \methodname{} (pos=FKL, neg=GRPO, 48.9) substantially outperforms all alternatives. FKL is well-suited to positive samples because their answers match the pseudo-label by definition: even when the label is wrong, the teacher is conditioned on the same answer the student produced, reducing to thinking-to-non-thinking distillation (45.7) that safely transfers careful reasoning. Hence positive-only FKL (46.7) outperforms all-FKL (46.3) and negative-only FKL (43.9), which forces the teacher to steer unmatched answers and injects corrupted signals. GRPO is better suited to negative samples: although its credit assignment is coarser, on hard TTT data most negatives are correctly identified (answer $\neq$ pseudo-label $\land$ $\neq$ ground truth), so a label-free penalty is strictly safer than corrupted distillation. GRPO on positives (37.2) lacks this robustness---it directly reinforces trajectories, and wrong pseudo-labels reverse the update with no mitigation. The reversed assignment (pos=GRPO, neg=FKL, 37.2) therefore performs worst, combining brittle reinforcement on positives with corrupted distillation on negatives.

\subsection{Analysis}
\label{sec:exp:analysis}

\paragraph{Privileged information.}

\begin{wraptable}{r}{0.50\textwidth}
\vspace{-13pt}
\centering
\small
\caption{Effect of privileged information type and teacher thinking mode (Qwen3-1.7B, OpenThoughts, AIME26). Columns: whether teacher uses thinking mode. Rows: privileged information injected into teacher prompt. Results in parentheses are evaluated with thinking disabled.}
\label{tab:privilege}
\begin{tabular}{l|cc}
\toprule
\textbf{Privilege} & \textbf{TM-on Teacher} & \textbf{TM-off Teacher} \\
\midrule
None & 45.8 (36.1) & 41.7 (8.1) \\
Answer & \cellcolor[HTML]{E4F0F9}\textbf{46.5} (\textbf{39.8}) & 33.6 (6.7) \\
Trajectory & 41.1 (8.9) & 40.8 (10.6) \\
\bottomrule
\end{tabular}
\vspace{-6pt}
\end{wraptable}

Table~\ref{tab:privilege} ablates privileged information under different teacher modes. With a thinking-mode teacher, the teacher--student distributional gap is already large, so a short answer suffices as a lightweight hint that steers the teacher without crowding out its reasoning. Even a wrong pseudo-label remains consistent with the student's answer in positive samples, and the update degenerates into thinking-to-non-thinking distillation---still a well-posed and beneficial signal (46.5 vs.\ 45.8). A full trajectory, by contrast, dominates the context and reduces the teacher to completing a given prefix rather than reasoning independently, degrading performance (41.1). With a non-thinking teacher, the gap is inherently small: a short answer barely shifts the distribution (33.6), while a full trajectory supplies needed context (40.8) but leaves both sides under weak reasoning, making training highly sensitive to pseudo-label noise. The thinking-teacher + answer setting thus offers the best trade-off between guidance and robustness.

\paragraph{Generalization beyond the target task.}

\begin{figure}[t]
\centering
\includegraphics[width=\textwidth]{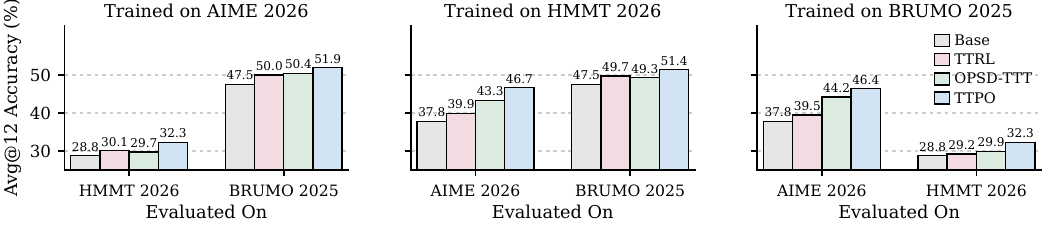}
\caption{Cross-benchmark generalization (Qwen3-1.7B). Each subplot corresponds to a training benchmark; each group within a subplot shows performance on a different target benchmark.}
\label{fig:generalization}
\end{figure}

To verify that \methodname{} acquires generalizable reasoning improvements rather than overfitting to specific problems, we train on each benchmark separately and evaluate on all three (Figure~\ref{fig:generalization}). Models trained on any single benchmark consistently improve on the other two as well. This cross-benchmark transfer confirms that \methodname{} strengthens underlying reasoning capabilities rather than memorizing problem-specific patterns.

\paragraph{Upper bound with labeled supervision.}

\begin{figure}[t]
\centering
\begin{minipage}[t]{0.4\textwidth}
\centering
\includegraphics[width=\textwidth]{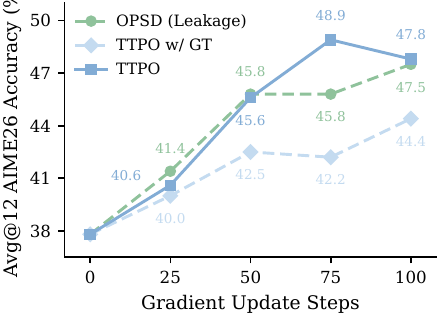}
\end{minipage}
\hfill
\begin{minipage}[t]{0.56\textwidth}
\centering
\includegraphics[width=\textwidth]{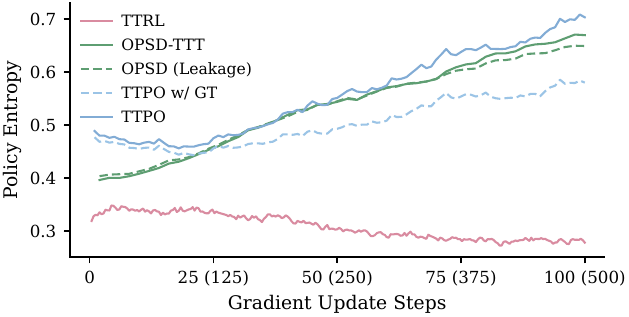}
\end{minipage}
\caption{\textbf{Left:} Comparison between pseudo-label vs.\ ground-truth supervision. \methodname{} w/ GT replaces majority-vote pseudo-labels with ground truth while keeping the asymmetric objective and token-level selection intact; OPSD (Leakage) trains standard OPSD directly on AIME26. \textbf{Right:} Entropy during training. Values in parentheses on the x-axis denote TTRL's training steps.}
\label{fig:upper_bound}
\end{figure}

We replace majority-vote pseudo-labels with ground-truth answers to probe the performance ceiling (Figure~\ref{fig:upper_bound}). Surprisingly, \methodname{} with pseudo-labels outperforms both \methodname{} w/ GT and OPSD (Leakage). First, perfectly correct labels are hard to match on difficult problems, yielding few or zero positives per instance; this starves the FKL branch and leaves GRPO with near-zero advantages that barely penalize negatives (Figure~\ref{fig:loss_curve}). Majority-vote labels, being easier to match, keep a healthy positive--negative split and both branches active. Second, AIME26 ground-truth answers are short numbers that barely shift the thinking teacher, unlike the richer OpenThoughts trajectories---reliable, yet too brief to guide strongly. The entropy plot (right) corroborates this: \methodname{} with pseudo-labels sustains higher entropy, as majority voting and distillation jointly promote exploration that compensates for---and ultimately surpasses---the theoretical benefit of perfect labels.

\paragraph{Sustainable self-evolution.} 

\begin{wrapfigure}{r}{0.4\textwidth}
\vspace{-12pt}
\centering
\includegraphics[width=0.4\textwidth]{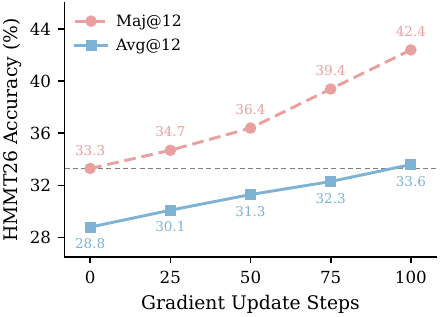}
\vspace{-8pt}
\caption{Avg@12 and Maj@12 during \methodname{} TTT on HMMT26 (Qwen3-1.7B). Dashed line: base Maj@12.}
\vspace{-9pt}
\label{fig:avg_maj}
\end{wrapfigure}

Since majority voting generates the training signal, the base model's Maj@12 sets the initial ceiling on pseudo-label quality. We track Avg@12 and Maj@12 throughout training to examine whether \methodname{} can break this ceiling (Figure~\ref{fig:avg_maj}). As training progresses, Avg@12 rises steadily to the base Maj@12, confirming that the collective knowledge in majority voting is distilled into single-sample performance. More importantly, Maj@12 does not stagnate but rises in tandem: as the model improves, higher-quality rollouts yield more accurate pseudo-labels, which in turn raise the training ceiling for later steps. This self-evolving cycle enables \methodname{} to improve beyond its initial supervision and ultimately outperform training with ground-truth that exceed the model's current capacity (Figure~\ref{fig:upper_bound}, left).

\section{Conclusion}
\label{sec:conclusion}

We introduced \methodname{}, which brings OPSD into label-free test-time training by combining it with RL under an asymmetric objective. Distillation provides dense token-level guidance on positives, while RL supplies a robust, label-free penalty on negatives. With token-level selection on both branches, \methodname{} matches ground-truth-supervised methods, substantially outperforms existing label-free approaches, and exhibits self-evolution with strong cross-task generalization.

\bibliography{iclr2027_conference}
\bibliographystyle{iclr2027_conference}

\appendix
\section{Implementation Details}
\label{app:impl}

We provide complete training and evaluation configurations in Tables~\ref{tab:config_all} and~\ref{tab:config_eval}. All experiments use the AdamW~\citep{loshchilov2017decoupled} optimizer with bfloat16 precision and Flash Attention 2. \methodname{} experiments use 4$\times$H20 GPUs; all other methods use 8$\times$H20 GPUs. We adopt full-vocabulary logit distillation for all distillation-based methods (OPSD, OPSD-TTT, and \methodname{}). Following~\citet{zhao2026self}, we use a thinking-mode-off student / thinking-mode-on teacher configuration, and the teacher is fixed to base model weights (LoRA adapters disabled) throughout training. For \methodname{} and OPSD-TTT, we set the maximum sampling length to 16,000 tokens to reduce answer extraction failures caused by truncation during majority voting, while the gradient update still only applies to the first 1,024 completion tokens.

\begin{table}[h]
\centering
\small
\caption{Training configuration for all methods. GRPO and OPSD use ground-truth labels; TTRL, OPSD-TTT, and \methodname{} are label-free. ``--'' indicates not applicable.}
\label{tab:config_all}
\setlength{\tabcolsep}{5pt}
\begin{tabular}{lccc}
\toprule
\textbf{Parameter} & \textbf{GRPO / TTRL} & \textbf{OPSD / OPSD-TTT} & \textbf{\methodname{}} \\
\midrule
\multicolumn{4}{l}{\textit{General}} \\
Learning Rate & $5 \times 10^{-6}$ & $5 \times 10^{-6}$ & $5 \times 10^{-6}$ \\
Max Gradient Norm & -- & 0.1 & 0.1 \\
Effective Batch Size & 32 & 32 & 32 \\
Training Steps & 500 & 100 & 100 \\
\midrule
\multicolumn{4}{l}{\textit{LoRA Configuration}} \\
LoRA Rank ($r$) & 64 & 64 & 64 \\
LoRA Alpha ($\alpha$) & 128 & 128 & 128 \\
Target Modules & \multicolumn{3}{c}{q\_proj, k\_proj, v\_proj, o\_proj, gate\_proj, up\_proj, down\_proj} \\
\midrule
\multicolumn{4}{l}{\textit{Generation}} \\
Number of Train Rollouts & 8 & 1 & 8 \\
Max Gradient Tokens & 16,000 & 1,024 & 1,024 \\
Sampling Temperature & 1.2 & 1.1 & 1.1 \\
Top-$p$ & -- & 0.95 & 0.95 \\
Top-$k$ & -- & 20 & 20 \\
KL Coefficient ($\beta$) & 0.0 & -- & -- \\
JSD Token Clip ($\tau$) & -- & 0.05 (1.7B, 4B) / 0.06 (8B) & 0.05 (1.7B, 4B) / 0.06 (8B) \\
\bottomrule
\end{tabular}
\end{table}

\begin{table}[h]
\centering
\caption{Evaluation configuration.}
\label{tab:config_eval}
\begin{tabular}{ll}
\toprule
\textbf{Parameter} & \textbf{Value} \\
\midrule
Thinking Mode & Enabled \\
Samples per Prompt & 12 \\
Temperature & 1.0 \\
Top-$p$ & 0.95 \\
Max New Tokens & 38,912 \\
Metric & Avg@12 \\
\bottomrule
\end{tabular}
\end{table}

\section{Prompt Templates}
\label{app:prompts}

We list the prompt templates used for both the student and teacher. All prompts are wrapped with the model's chat template via {\small\textsf{apply\_chat\_template}}. When \textsf{privilege\_info = none} (the no-privilege ablation in Table~\ref{tab:privilege}), the teacher prompt is identical to the student prompt but with thinking mode enabled.

\begin{promptbox}[Student Prompt (thinking mode off)]
Problem: \{problem\}

Please reason step by step, and put your final answer within \textbackslash boxed\{\}.
\end{promptbox}

\begin{promptbox}[Teacher Prompt (thinking mode on, privilege = answer)]
Problem: \{problem\}

Here is a reference solution to this problem:\\
=== Reference Solution Begin ===\\
\{pseudo\_label\}\\
=== Reference Solution End ===

After reading the reference solution above, make sure you truly understand the reasoning behind each step --- do not copy or paraphrase it. Now, using your own words and independent reasoning, derive the same final answer to the problem above. Think step by step, explore different approaches, and don't be afraid to backtrack or reconsider if something doesn't work out:

Please reason step by step, and put your final answer within \textbackslash boxed\{\}.
\end{promptbox}

\section{Case Study: Token-Level Selection Visualization}
\label{app:token_viz}

We visualize how our token weighting (\S\ref{sec:method:positive}) and token masking (\S\ref{sec:method:negative}) operate on a concrete example. Excerpts are drawn from a positive sample and a negative sample generated for the same geometry problem.\footnote{Problem: Given square $ABCD$ with $L$ on $CD$, $K$ on extension of $DA$ beyond $A$, $\angle KBL=90^\circ$, $KD=19$, $CL=6$. Find $LD$. Ground truth: $7$.} The positive sample reaches the correct answer $7$; the negative arrives at $\sqrt{397}$. Recall that token weighting down-weights low-entropy, low-divergence positions where the student has converged, while token masking suppresses low-probability, low-entropy positions representing anomalous model outputs.

\subsection{Token Weighting}

\begin{AIbox}[colback=posboxbg,colframe=posboxframe,colbacktitle=posboxtitle]{Positive Sample --- Token Weighting (Answer: 7 \checkmark)}
\small\ttfamily
\textbf{Legend:} \tw{5}{\,}\tw{15}{\,}\tw{25}{\,}\tw{35}{\,}\tw{45}{\,}\tw{55}{\,}\tw{65}{\,}\tw{75}{\,}\tw{85}{\,}\tw{95}{\,} low $\longrightarrow$ high weight

\medskip
\noindent\textbf{Segment: Geometric reasoning to determine $K$'s coordinates}

\medskip
\tw{90}{Since} \$\tw{30}{DA}\$ \tw{30}{goes} from \$D = (0, a)\$ to \$A = (0, 0)\$,\tw{70}{ the}\tw{70}{ extension}\tw{15}{ beyond} \$A\$\tw{70}{ is}\tw{70}{ the}\tw{50}{ line} \tw{30}{\$}x = 0\tw{30}{\$,}\tw{90}{ going}\tw{90}{ downwards}\tw{90}{.}\tw{30}{ So}\tw{50}{ \$}K\tw{70}{\$}\tw{70}{ is}\tw{70}{ at}\tw{30}{ \$(}0, \tw{30}{y}\$)\tw{70}{,}\tw{70}{ for} some \$y < 0\$.
\end{AIbox}

The contrast is stark: coordinate values (``$(0, a)$'', ``$(0, 0)$'', ``$x = 0$'') receive near-zero weight---these are deterministic once the setup is chosen, and both student and teacher assign near-unit probability to each digit. High weight concentrates on geometric \emph{insights} that determine the solution strategy: ``the extension beyond $A$ is the line... going downwards'' (identifying the geometric locus) and ``So $K$ is at $(0, y)$, for some $y < 0$'' (drawing the conclusion). These are positions where the student is uncertain which geometric fact to invoke or the teacher favors a different continuation, making them the sole source of learning signal. Token weighting thus focuses distillation on \textbf{where to reason and what to conclude}, not on reproducing mechanical substitutions the model already handles reliably.

\subsection{Token Masking}

\begin{AIbox}[colback=negboxbg,colframe=negboxframe,colbacktitle=negboxtitle]{Negative Sample --- Token Masking (Answer: $\sqrt{397}$ \ding{55})}
\small\ttfamily
\textbf{Legend:} kept token, \mk{masked token}

\medskip
\noindent\textbf{Segment: Erroneous coordinate derivation (root cause of failure)}

\medskip
\$CL\mk{ = 6}\$: Since \$C =\mk{ (s, s)}\$ and \$L\$ is on \mk{\$CD\$,} we move\mk{ 6} units from \$\mk{C}\$ along \$\mk{CD}\$ (which is vertical\mk{).}

So, \mk{point} \$ \mk{L} = \mk{(s, s} - \mk{6})\$
\end{AIbox}

The kept tokens are precisely the positions worth penalizing: \emph{local errors} (``which is vertical''---$CD$ is actually horizontal under the model's own coordinates), \emph{context-inconsistent expressions} (``we move... units from $C$ along $CD$''---applying a vertical displacement to a horizontal segment), and the resulting \emph{anomalous conclusion} (``So, $L=$''---committing to coordinates that place $L$ off the intended side). Masked-out tokens, by contrast, are locally correct arithmetic (``$(s, s)$'', ``$= 6$'') and formatting that would appear identically in a correct solution; penalizing them would damage valid computation skills without addressing the actual reasoning flaw. Token masking thus restricts the penalty gradient to the \textbf{confident errors that cause failure}---wrong geometric claims and their immediate consequences---while leaving shared, reusable sub-skills intact.

\subsection{Complementarity}

The two mechanisms implement a dual philosophy: token weighting asks ``\emph{where does the model still need to learn?}'' and suppresses already-converged positions in positive samples; token masking asks ``\emph{where is the model confidently wrong?}'' and suppresses locally-correct or uncertain positions in negative samples. Both avoid wasting gradient on low-signal tokens---from opposite directions---yielding more efficient and stable training.

\section{Additional Experiments}
\subsection{Training Dynamics}
\label{app:training_dynamics}

\begin{wrapfigure}{r}{0.56\textwidth}
\centering
\vspace{-12pt}
\includegraphics[width=0.56\textwidth]{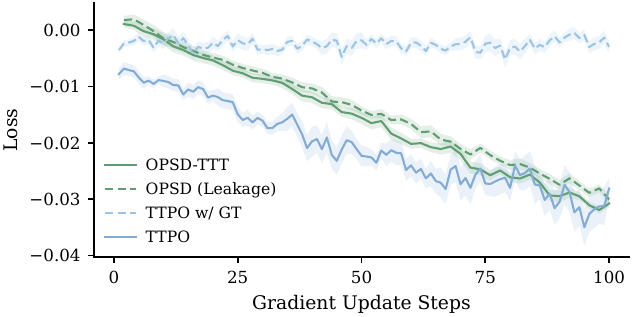}
\vspace{-10pt}
\caption{Training loss curves.}
\label{fig:all_loss}
\vspace{-8pt}
\end{wrapfigure}

Figure~\ref{fig:all_loss} compares training dynamics of Qwen3-1.7B TTT on AIME 2026 across methods. The most striking observation is that \methodname{} w/ GT exhibits dramatically weaker training signal than all other methods: its loss barely decreases and frequently stagnates near zero. This directly validates our theoretical analysis (Eq.~\ref{eq:gt_fkl}--\ref{eq:gt_grpo}) — on hard AIME problems where $|\mathcal{P}_{\text{GT}}| \approx 0$, GT routing starves both branches simultaneously. In contrast, \methodname{} with majority-vote pseudo-labels maintains a steady and substantial loss decrease throughout training, confirming that vote-based routing keeps both branches active. OPSD (Leakage) and OPSD-TTT both show consistent loss reduction; notably, \methodname{} achieves comparable or stronger training dynamics despite operating in a fully label-free setting.

\subsection{Non-Thinking Evaluation}
\label{app:nonthinking}

\begin{table}[h]
\centering
\small
\caption{Non-thinking evaluation on OpenThoughts training data. Models are evaluated with thinking mode \textbf{disabled}. Both OPSD and \methodname{} are trained with a thinking-mode-on teacher.}
\label{tab:nonthinking}
\begin{tabular}{p{2.4cm}|cccccc}
\toprule
\textbf{Method} & \textbf{AIME25} & \textbf{HMMT25} & \textbf{AIME26} & \textbf{HMMT26} & \textbf{BRUMO25} & \textbf{Average} \\
\midrule
\multicolumn{7}{@{}l}{\textit{Qwen3-1.7B}} \\
\hspace{0.8em}Base & 9.2 & 5.6 & 8.8 & 6.1 & 17.8 & 9.5 \\
\hspace{0.8em}+OPSD$^\dagger$ & \underline{16.9} & \underline{9.2} & \underline{19.4} & \underline{11.9} & \underline{25.6} & \underline{16.6} \\
\rowcolor[HTML]{E4F0F9} \hspace{0.8em}+\methodname{} & \textbf{39.2} & \textbf{20.6} & \textbf{39.8} & \textbf{26.3} & \textbf{47.5} & \textbf{34.7} \\
\midrule
\multicolumn{7}{@{}l}{\textit{Qwen3-4B}} \\
\hspace{0.8em}Base & 22.2 & 12.5 & 19.4 & 17.2 & 28.3 & 19.9 \\
\hspace{0.8em}+OPSD$^\dagger$ & \underline{26.7} & \underline{18.9} & \underline{24.4} & \underline{22.5} & \underline{36.1} & \underline{25.7} \\
\rowcolor[HTML]{E4F0F9} \hspace{0.8em}+\methodname{} & \textbf{57.2} & \textbf{36.7} & \textbf{61.4} & \textbf{36.4} & \textbf{60.8} & \textbf{50.5} \\
\midrule
\multicolumn{7}{@{}l}{\textit{Qwen3-8B}} \\
\hspace{0.8em}Base & 20.6 & 11.4 & 21.1 & 18.7 & 29.7 & 20.3 \\
\hspace{0.8em}+OPSD$^\dagger$ & \underline{25.0} & \underline{14.2} & \underline{22.2} & \underline{19.9} & \underline{37.8} & \underline{23.8} \\
\rowcolor[HTML]{E4F0F9} \hspace{0.8em}+\methodname{} & \textbf{67.8} & \textbf{42.5} & \textbf{65.0} & \textbf{41.2} & \textbf{67.2} & \textbf{56.7} \\
\bottomrule
\end{tabular}
\end{table}

Table~\ref{tab:nonthinking} evaluates models with thinking mode disabled to assess whether training with a thinking-mode teacher transfers reasoning capabilities to non-thinking inference. \methodname{} achieves dramatically larger gains than OPSD across all scales: on average, \methodname{} improves over the base model by +25.2 (1.7B), +30.6 (4B), and +36.4 (8B) points, while OPSD improves by only +7.1, +5.8, and +3.5 points respectively. This indicates that \methodname{} far more effectively absorbs the thinking teacher's reasoning ability into the student's non-thinking distribution. We attribute this to the asymmetric objective: the GRPO branch on negative samples directly penalizes poor reasoning patterns in the student's own generation mode, while OPSD's pure distillation only passively aligns the student toward the teacher without actively suppressing failure modes.

\subsection{Additional Ablations}
\label{app:ablation}

\paragraph{RL weight $\lambda$.}

\begin{figure}[h]
\centering
\includegraphics[width=\textwidth]{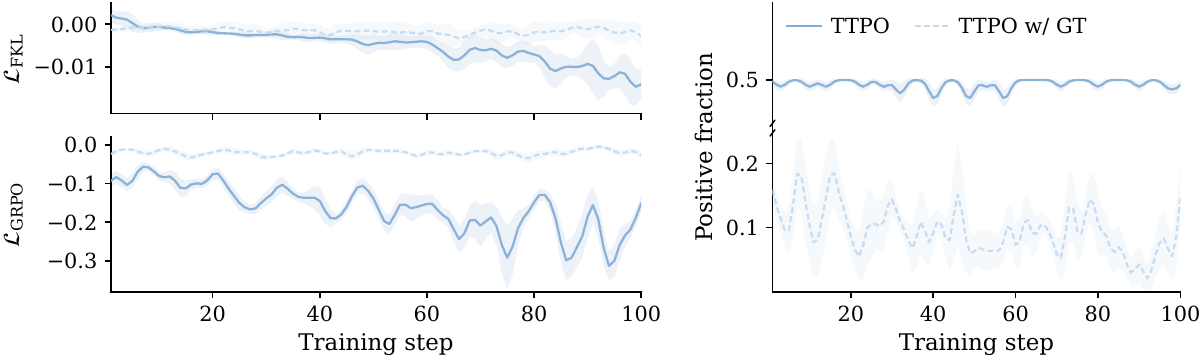}
\caption{\textbf{Left:} Training loss curves for the OPSD and GRPO branches (unweighted). \textbf{Right:} Positive sample fraction over training steps. (Qwen3-1.7B, OpenThoughts).}
\label{fig:loss_curve}
\end{figure}

\begin{table}[h]
\centering
\small
\caption{Ablation on RL weight $\lambda$ (Qwen3-1.7B, OpenThoughts).}
\label{tab:rl_weight}
\begin{tabular}{c|ccc}
\toprule
$\lambda$ & \textbf{AIME26} & \textbf{HMMT26} & \textbf{BRUMO25} \\
\midrule
0.01 & 41.4 & 28.8 & 51.1 \\
0.05 & 43.9 & 29.0 & 50.6 \\
\rowcolor[HTML]{E4F0F9} 0.10 & \textbf{46.5} & \textbf{31.6} & \textbf{54.7} \\
0.15 & 44.7 & 29.5 & 50.3 \\
0.20 & 41.7 & 26.3 & 51.6 \\
\bottomrule
\end{tabular}
\end{table}

As shown in Figure~\ref{fig:loss_curve}, the raw GRPO loss is roughly an order of magnitude larger than the OPSD forward-KL loss, while the positive sample fraction increases over training as the model produces more correct answers. Without scaling, the GRPO branch would dominate gradients and destabilize training. We therefore introduce a weight $\lambda$ on the GRPO loss to balance the two branches. Table~\ref{tab:rl_weight} sweeps $\lambda \in \{0.01, 0.05, 0.1, 0.15, 0.2\}$: performance peaks at $\lambda{=}0.1$, which approximately equalizes the gradient magnitudes of the two branches. Both under-weighting ($\lambda{\leq}0.05$, insufficient negative penalty) and over-weighting ($\lambda{\geq}0.15$, excessive penalty dominating distillation) degrade results.

\paragraph{$K_{\text{train}}$ positive-negative fraction.}

\begin{table}[h]
\begin{minipage}[t]{0.48\textwidth}
\centering
\small
\caption{Positive-negative fraction in $K_{\text{train}}$ (Qwen3-1.7B, OpenThoughts). Dynamic: ratio mirrors full $K$ rollouts.}
\label{tab:fraction}
\resizebox{\textwidth}{!}{%
\begin{tabular}{l|ccc}
\toprule
\textbf{Fraction} & \textbf{AIME26} & \textbf{HMMT26} & \textbf{BRUMO25} \\
\midrule
Random & 45.8 & 30.3 & 50.9 \\
\rowcolor[HTML]{E4F0F9} Fixed (0.5) & \textbf{46.5} & \textbf{31.6} & \textbf{54.7} \\
Dynamic & 46.1 & 31.1 & 51.1 \\
\bottomrule
\end{tabular}}
\end{minipage}
\hfill
\begin{minipage}[t]{0.48\textwidth}
\centering
\small
\caption{$K_{\text{train}}$ selection strategy (Qwen3-1.7B, OpenThoughts).}
\label{tab:selection}
\resizebox{\textwidth}{!}{%
\begin{tabular}{l|ccc}
\toprule
\textbf{Strategy} & \textbf{AIME26} & \textbf{HMMT26} & \textbf{BRUMO25} \\
\midrule
Random & 45.0 & 31.1 & 50.7 \\
\rowcolor[HTML]{E4F0F9} Shortest & \textbf{46.5} & \textbf{31.6} & \textbf{54.7} \\
Longest & 45.6 & 30.8 & 51.1 \\
Top signal & 45.8 & 30.6 & 51.9 \\
\bottomrule
\end{tabular}}
\end{minipage}
\end{table}

Table~\ref{tab:fraction} ablates the positive-negative composition of the $K_{\text{train}}$ subset. A fixed 50/50 split outperforms both random sampling and a dynamic fraction. The dynamic strategy faces a fundamental dilemma: when the positive fraction in $K$ is high (i.e., pseudo-label is likely correct), proportionally reducing negative samples in $K_{\text{train}}$ cancels the amplified group-relative advantages that negative samples receive — the enlarged signal is immediately diluted by fewer recipients. Conversely, if the dynamic strategy inverts the ratio (more negatives when positives dominate), it sacrifices the reliable fine-grained FKL supervision available precisely when the pseudo-label is most trustworthy, replacing it with coarser GRPO penalties. Either direction has drawbacks; a fixed 50/50 split avoids both failure modes and guarantees stable gradient contributions from both branches at every step.

\paragraph{$K_{\text{train}}$ selection strategy.}
Table~\ref{tab:selection} compares strategies for selecting which rollouts enter $K_{\text{train}}$. Selecting the shortest completions performs best. Since only the first 1,024 tokens participate in the gradient update, shorter trajectories ensure that these tokens constitute a larger fraction of the total reasoning chain and are more likely to contain the critical steps that determine the final answer. For longer trajectories, the first 1,024 tokens often cover only preliminary exploration, with the decisive reasoning occurring well beyond the gradient window — yielding little useful learning signal. The ``Top signal'' strategy selects positive samples with the highest teacher-student FKL divergence (intuitively, trajectories where the student deviates most from the teacher and thus has the most to learn) and negative samples with the highest log-probability (intuitively, confident errors that carry the strongest penalty signal). Despite this seemingly stronger per-sample signal, the strategy underperforms shortest selection — the intuition that larger divergence or higher confidence implies more useful gradients lacks theoretical grounding and appears unreliable in practice.

\section{Theoretical Analysis}
\label{app:theory}

\subsection{Setup}

For problem $x$ with ground-truth $a^*$ and pseudo-label $\hat{a}$:
\begin{align}
    q_t^{(a)} = \pi_\theta(\cdot \mid [x; a], y_{<t}), \quad p_t = \pi_\theta(\cdot \mid x, y_{<t}), \quad
    \nabla_\theta \mathrm{KL}(q_t^{(a)} \| p_t) = -\sum_{v} q_t^{(a)}(v)\, \nabla_\theta \log p_t(v).
    \label{eq:fkl_grad}
\end{align}

\subsection{FKL Signal Analysis}

\paragraph{Positive samples ($y_k \in \mathcal{P}$, $a_k = \hat{a}$).}

The trajectory agrees with teacher, so the FKL collapses to:
\begin{equation}
    \mathrm{KL}\!\left(q_t^{(\hat{a})} \,\big\|\, p_t\right)\bigg|_{a_k = \hat{a}} \;=\; \underbrace{\mathrm{KL}(q_t \| p_t)}_{\text{thinking vs.\ non-thinking only}}.
    \label{eq:pos_fkl}
\end{equation}
If $\hat{a} = a^*$ this recovers standard OPSD; if $\hat{a} \neq a^*$ it reduces to answer-agnostic distillation.

\paragraph{Negative samples ($y_k \in \mathcal{N}$, $a_k \neq \hat{a}$).}

The disagreement introduces a non-negative conflict term:
\begin{equation}
    \mathrm{KL}\!\left(q_t^{(\hat{a})} \,\big\|\, p_t\right)\bigg|_{a_k \neq \hat{a}} \;=\; \underbrace{\mathrm{KL}(q_t \| p_t)}_{\text{thinking vs.\ non-thinking}} \;+\; \underbrace{\Delta_{\text{conflict}}(t)}_{\geq\, 0},
    \label{eq:neg_fkl}
\end{equation}
where $\Delta_{\text{conflict}}(t)$ captures the teacher's pressure to redirect reasoning from $a_k$ toward $\hat{a}$. This is beneficial when $\hat{a} = a^*$, but harmful when $\hat{a} \neq a^*$, particularly when $a_k = a^*$, as the gradient actively suppresses correct reasoning.

\paragraph{Method comparison.}

By Eq.~\ref{eq:pos_fkl}, FKL on $\mathcal{P}$ is at worst benign and at best recovers standard OPSD; GRPO on $\mathcal{N}$ is label-agnostic. U-OPSD applies FKL to $\mathcal{N}$ instead, where $\Delta_{\text{conflict}}$ (Eq.~\ref{eq:neg_fkl}) misdirects correct trajectories whenever $\hat{a} \neq a^*$, and requires reliable pseudo-labels since low consensus makes $\Delta_{\text{conflict}}$ harmful while unanimous consensus yields waste of rollouts ($|\mathcal{N}|=0$).

\subsection{Majority-Vote vs.\ Ground-Truth Routing}

With binary reward $r_k = \mathbf{1}[a_k \in \text{majority}]$, $\bar{r} = |\mathcal{P}|/K$. The GRPO advantage for $k \in \mathcal{N}$:
\begin{equation}
    A_k = -\sqrt{\frac{|\mathcal{P}|/K}{1 - |\mathcal{P}|/K}}, \quad k \in \mathcal{N}.
    \label{eq:advantage}
\end{equation}

\paragraph{GT routing} ($\mathcal{P}_{\text{GT}} = \{k: a_k = a^*\}$).
On hard problems with small $|\mathcal{P}_{\text{GT}}|$, both branches vanish:
\begin{align}
    \frac{|\mathcal{P}_{\text{GT}}|}{K} \approx 0
    \quad&\text{and}\quad
    \underbrace{\sum_{k \in \mathcal{P}_{\text{GT}}} \nabla_\theta \mathrm{KL}(q_t \| p_t)}_{\text{FKL}} \approx \mathbf{0}, \label{eq:gt_fkl}\\
    |A_k| = \sqrt{\frac{|\mathcal{P}_{\text{GT}}|/K}{1 - |\mathcal{P}_{\text{GT}}|/K}} \approx 0
    \quad&\text{and}\quad
    \underbrace{A_k \,\nabla_\theta \log \pi_\theta(y_k \mid x)}_{\text{GRPO}} \approx \mathbf{0}. \label{eq:gt_grpo}
\end{align}

\paragraph{Vote routing} ($\mathcal{P}_{\text{vote}} = \{k: a_k = \hat{a}\}$).
The model always forms a non-empty consensus $|\mathcal{P}_{\text{vote}}| > 0$, so both branches remain active:
\begin{align}
    \frac{|\mathcal{P}_{\text{vote}}|}{K} > 0
    \quad&\text{and}\quad
    \underbrace{\sum_{k \in \mathcal{P}_{\text{vote}}} \nabla_\theta \mathrm{KL}(q_t \| p_t)}_{\text{FKL}} \neq \mathbf{0}, \label{eq:vote_fkl}\\
    A_k = -\sqrt{\frac{|\mathcal{P}_{\text{vote}}|/K}{1 - |\mathcal{P}_{\text{vote}}|/K}} < 0
    \quad&\text{and}\quad
    \underbrace{A_k \,\nabla_\theta \log \pi_\theta(y_k \mid x)}_{\text{GRPO}} \neq \mathbf{0}. \label{eq:vote_grpo}
\end{align}
As the model improves, $\hat{a}$ converges toward $a^*$ (Figure~\ref{fig:avg_maj}).

\section{Limitations and Future Work}
\label{app:limitations}

\paragraph{Dependence on majority-vote quality.} \methodname{} relies on majority voting to generate pseudo-labels and classify positive/negative samples. When the sample budget $K$ is very small or the problem is so difficult that no rollout produces a correct answer, the voting signal degrades and both branches receive noisy supervision. Adaptive strategies that adjust the positive-negative ratio or fall back to pure RL under low-consensus conditions could mitigate this.

\paragraph{Domain scope.} Our experiments are restricted to mathematical reasoning with verifiable final answers. Extending \methodname{} to domains where correctness is harder to extract automatically --- such as code generation (requiring execution-based verification) or open-ended reasoning (requiring learned reward models) --- remains unexplored.

\paragraph{Dynamic training curriculum.} \methodname{} applies a fixed asymmetric objective throughout training. As the model improves and pseudo-label accuracy rises, the optimal balance between distillation and RL may shift. A curriculum that dynamically adjusts the RL weight or positive-negative fraction to training dynamics could further improve efficiency.

\end{document}

%% file: math_commands.tex
\usepackage{amsmath,amsfonts,bm}

\def\eqref#1{equation~\ref{#1}}

\def\1{\bm{1}}

\DeclareMathAlphabet{\mathsfit}{\encodingdefault}{\sfdefault}{m}{sl}
\SetMathAlphabet{\mathsfit}{bold}{\encodingdefault}{\sfdefault}{bx}{n}

%% file: figures/style.tex
\usetikzlibrary{patterns,patterns.meta,positioning,backgrounds,calc}

\definecolor{mila-purple}{RGB}{102,46,125}
\definecolor{mila-khaki}{RGB}{245,231,206}
\definecolor{mila-khaki-dark}{RGB}{185,174,154}
\definecolor{mila-yellow}{HTML}{F9B40D}
\definecolor{mila-yellow-dark}{HTML}{DD950B}
\definecolor{mila-greenblue}{RGB}{130,207,190}
\definecolor{plot-blue}{RGB}{132,182,206}
\definecolor{plot-red}{RGB}{248,191,199}
\definecolor{plot-green}{RGB}{157,207,127}
\definecolor{mila-purple-1}{HTML}{EFE8F1}
\definecolor{mila-purple-2}{HTML}{D8C9DD}
\definecolor{mila-purple-3}{HTML}{C1AACA}
\definecolor{mila-purple-4}{HTML}{AA8BB7}
\definecolor{mila-purple-5}{HTML}{936BA3}
\definecolor{mila-purple-6}{HTML}{7C4D90}
\definecolor{mila-purple-dark}{HTML}{451059}
\definecolor{mila-purple-superdark}{HTML}{340040}
\definecolor{delethink-blue}{HTML}{3ba1ff}
\definecolor{delethink-purple}{HTML}{6441D2}
\definecolor{delethink-dark-purple}{HTML}{3C2880}

\definecolor{inftythink-red}{RGB}{248,191,199}
\definecolor{inftythink-blue}{RGB}{150,184,243}
\definecolor{inftythink-green}{RGB}{154,226,192}
\definecolor{inftythink-yellow}{RGB}{249,180,13}

\newlength{\sz}         \setlength{\sz}{0.7cm}
\newlength{\sw}         \setlength{\sw}{0.4cm}
\newlength{\sww}        \setlength{\sww}{0.37cm}
\newlength{\rectr}         \setlength{\rectr}{5pt}
\newlength{\rectw}         \setlength{\rectw}{2cm}
\newlength{\rectwconclu}   \setlength{\rectwconclu}{1cm}
\newlength{\rectwlongcot}  \setlength{\rectwlongcot}{15cm}
\newlength{\segap}      \setlength{\segap}{1.2cm}
\newlength{\segaparrow} \setlength{\segaparrow}{0.1cm}

\newlength{\BORDER}
\setlength{\BORDER}{1pt}

\tikzset{
  border/.style={line width=\BORDER},
  centerline/.style={dash pattern=on 1pt off 2pt, line cap=round, line width=\BORDER},
  arrowline/.style={->, line width=\BORDER},
  curvedarrow/.style={->, draw=black!30, line width=0.7pt, shorten >=2pt, shorten <=4pt}
}

\tikzset{
  segment marker shape/.is choice,
  segment marker shape/circle/.code={%
    \def\drawmarker##1{\fill[white] (##1) circle[radius=1.2pt];}%
  },
  segment marker shape/square/.code={%
    \def\drawmarker##1{\fill[white]
      ($(##1)+(-1.1pt,-1.1pt)$) rectangle ($(##1)+(1.1pt,1.1pt)$);}%
  },
  segment marker shape/diamond/.code={%
    \def\drawmarker##1{\fill[white]
      ($(##1)+(0,1.4pt)$)--($(##1)+(1.4pt,0)$)--($(##1)+(0,-1.4pt)$)--($(##1)+(-1.4pt,0)$)--cycle;}%
  },
  segment marker shape/triangle/.code={%
    \def\drawmarker##1{\fill[white]
      ($(##1)+(-1.2pt,-1.2pt)$)--($(##1)+(-1.2pt,1.2pt)$)--($(##1)+(1.6pt,0)$)--cycle;}%
  },
  segment marker shape/trangle/.style = {segment marker shape/triangle},
}

\tikzset{
  pics/inftythink_iter_1/.style n args={3}{
    code={
      \begin{scope}[node distance=0pt]
        \begin{scope}[local bounding box=sq]
          \path[
            draw=#1, line width=1.4\BORDER, dashed,
            preaction={fill=#1!5}, pattern color=black!30
          ]
          (\rectr,0) -- (\sw,0) -- (\sw,\sz) -- (\rectr,\sz)
          arc (90:180:\rectr) -- (0,\rectr) arc (180:270:\rectr) -- cycle;
        \end{scope}
        \node[font=\bfseries\large] at (sq.center) {$\mathrm{q}$};

        \begin{scope}[local bounding box=rect, xshift=3pt+\sw]
          \path[
            draw=#2, line width=1.4\BORDER,
            preaction={fill=#2!10}, pattern color=black!30
          ]
          (0.8\rectw,0) -- (0,0) -- (0,\sz) -- (0.8\rectw,\sz);
        \end{scope}

        \begin{scope}[xshift=3pt+\sw+0.8\rectw, local bounding box=summary]
          \path[
            draw=#3, line width=1.4\BORDER,
            preaction={fill=#3!10}, pattern color=black!30
          ]
          (0,\sz) -- (0.4\rectw,\sz) -- (0.4\rectw,0) -- (0,0);
        \end{scope}

        \begin{scope}[on background layer]
          \fill[#2!10] (rect.south west) rectangle (rect.north east);
          \fill[#3!10] (summary.south west) rectangle (summary.north east);
        \end{scope}

        \begin{scope}
          \clip (rect.south west) rectangle (rect.north east);
        
          \def\dia{1.8pt}
        
          \coordinate (gxW) at ($(rect.west)!0.19!(rect.east)$);
          \coordinate (gxE) at ($(rect.west)!0.81!(rect.east)$);
        
          \coordinate (gy1) at ($(rect.south)!0.75!(rect.north)$);
          \coordinate (p1)  at ($(gxW|-gy1)$);
          \fill[#2!60]
            ($(p1)+(-\dia,0)$) -- ($(p1)+(0,\dia)$) -- ($(p1)+(\dia,0)$) -- ($(p1)+(0,-\dia)$) -- cycle;
          \draw[#2!60, line width=\BORDER, line cap=round]
            ($(p1)+(\dia+0.6pt,0)$) -- ([xshift=-2pt]gxE|-gy1);
        
          \coordinate (gy2) at ($(rect.south)!0.50!(rect.north)$);
          \coordinate (p2)  at ($(gxW|-gy2)$);
          \fill[#2!60]
            ($(p2)+(-\dia,0)$) -- ($(p2)+(0,\dia)$) -- ($(p2)+(\dia,0)$) -- ($(p2)+(0,-\dia)$) -- cycle;
          \draw[#2!60, line width=\BORDER, line cap=round]
            ($(p2)+(\dia+0.6pt,0)$) -- ([xshift=-5pt]gxE|-gy2);
        
          \coordinate (gy3) at ($(rect.south)!0.25!(rect.north)$);
          \coordinate (p3)  at ($(gxW|-gy3)$);
          \fill[#2!60]
            ($(p3)+(-\dia,0)$) -- ($(p3)+(0,\dia)$) -- ($(p3)+(\dia,0)$) -- ($(p3)+(0,-\dia)$) -- cycle;
          \draw[#2!60, line width=\BORDER, line cap=round]
            ($(p3)+(\dia+0.6pt,0)$) -- (gxE|-gy3);
        \end{scope}

        \begin{scope}
          \clip (summary.south west) rectangle (summary.north east);
        
          \path coordinate (gxW) at ($(summary.west)!0.18!(summary.east)$);
          \path coordinate (gxE) at ($(summary.west)!0.75!(summary.east)$);
        
          \path coordinate (gy1) at ($(summary.south)!0.75!(summary.north)$);
          \fill[#3!60] (gxW|-gy1) circle[radius=1.2pt];
          \draw[#3!60, line width=\BORDER, line cap=round]
               ($(gxW|-gy1)+(1.6pt,0)$) -- ([xshift=-0pt]gxE|-gy1);
        
          \path coordinate (gy2) at ($(summary.south)!0.50!(summary.north)$);
          \fill[#3!60] (gxW|-gy2) circle[radius=1.2pt];
          \draw[#3!60, line width=\BORDER, line cap=round]
               ($(gxW|-gy2)+(1.6pt,0)$) -- ([xshift=-3pt]gxE|-gy2);
        
          \path coordinate (gy3) at ($(summary.south)!0.25!(summary.north)$);
          \fill[#3!60] (gxW|-gy3) circle[radius=1.2pt];
          \draw[#3!60, line width=\BORDER, line cap=round]
               ($(gxW|-gy3)+(1.6pt,0)$) -- ([xshift=-6pt]gxE|-gy3);
        \end{scope}

        \coordinate (-sqwest)   at (sq.west);
        \coordinate (-rectwest) at (rect.west);
        \coordinate (-recteast) at (rect.east);
        \coordinate (-east)     at (summary.east);
        \coordinate (-north) at ([xshift=1pt]summary.north);
        \coordinate (-rectnorth) at ([xshift=0.25\rectw]rect.north);
        \coordinate (-rectsouth) at (rect.south);
      \end{scope}
    }
  },
  pics/inftythink_iter_1/.default={mila-purple-1}{mila-purple-2}
}

\tikzset{
  pics/inftythink_iter_n-1/.style n args={3}{
    code={
      \begin{scope}[node distance=0pt]
        \begin{scope}[local bounding box=sq]
          \path[draw=none, preaction={fill=#1!10}, pattern color=black!30]
            (\rectr,0) -- (\sww,0) -- (\sww,\sz) -- (\rectr,\sz)
            arc (90:180:\rectr) -- (0,\rectr) arc (180:270:\rectr) -- cycle;
          \draw[#1, dashed, line width=1.4\BORDER]
            (\sww,\sz) -- (\rectr,\sz) arc (90:180:\rectr) -- (0,\rectr)
            arc (180:270:\rectr) -- (\rectr,0) -- (\sww,0);
          \path[use as bounding box] (0,0) rectangle (\sww,\sz);
        \end{scope}
        \node[font=\bfseries\large] at (sq.center) {$\mathrm{q}$};

        \node[
          draw=none, minimum height=\sz, minimum width=0.4\rectw,
          inner sep=0pt, outer sep=0pt, right=0pt of sq
        ] (rectprmpt) {};
        \begin{scope}[on background layer]
          \fill[#3!10] (rectprmpt.south west) rectangle (rectprmpt.north east);
        \end{scope}
        \draw[#3, line width=1.4\BORDER, dashed]
          (rectprmpt.north west) -- (rectprmpt.north east)
          (rectprmpt.north east) -- (rectprmpt.south east)
          (rectprmpt.south east) -- (rectprmpt.south west);

        \begin{scope}
          \clip (rectprmpt.south west) rectangle (rectprmpt.north east);
        
          \path coordinate (gxW) at ($(rectprmpt.west)!0.19!(rectprmpt.east)$);
          \path coordinate (gxE) at ($(rectprmpt.west)!0.81!(rectprmpt.east)$);
        
          \path coordinate (gy1) at ($(rectprmpt.south)!0.75!(rectprmpt.north)$);
          \fill[#3!60] (gxW|-gy1) circle[radius=1.2pt];
          \draw[#3!60, line width=\BORDER, line cap=round]
               ($(gxW|-gy1)+(1.6pt,0)$) -- ([xshift=-0pt]gxE|-gy1);
        
          \path coordinate (gy2) at ($(rectprmpt.south)!0.50!(rectprmpt.north)$);
          \fill[#3!60] (gxW|-gy2) circle[radius=1.2pt];
          \draw[#3!60, line width=\BORDER, line cap=round]
               ($(gxW|-gy2)+(1.6pt,0)$) -- ([xshift=-3pt]gxE|-gy2);
        
          \path coordinate (gy3) at ($(rectprmpt.south)!0.25!(rectprmpt.north)$);
          \fill[#3!60] (gxW|-gy3) circle[radius=1.2pt];
          \draw[#3!60, line width=\BORDER, line cap=round]
               ($(gxW|-gy3)+(1.6pt,0)$) -- ([xshift=-6pt]gxE|-gy3);
        \end{scope}

        \begin{scope}[local bounding box=rect, xshift=3pt+\sw+0.4\rectw]
          \path[
            draw=#2, line width=1.4\BORDER,
            preaction={fill=#2!10}, pattern color=black!30
          ]
          (0.8\rectw,0) -- (0,0) -- (0,\sz) -- (0.8\rectw,\sz);
        \end{scope}

        \begin{scope}[xshift=3pt+\sw+1.2\rectw, local bounding box=summary]
          \path[
            draw=#3, line width=1.4\BORDER,
            preaction={fill=#3!10}, pattern color=black!30
          ]
          (0,\sz) -- (0.4\rectw,\sz) -- (0.4\rectw,0) -- (0,0);
        \end{scope}

        \begin{scope}[on background layer]
          \fill[#2!10] (rect.south west) rectangle (rect.north east);
          \fill[#3!10] (summary.south west) rectangle (summary.north east);
        \end{scope}
        
        \begin{scope}
          \clip (rect.south west) rectangle (rect.north east);
        
          \def\dia{1.8pt}
        
          \coordinate (gxW) at ($(rect.west)!0.19!(rect.east)$);
          \coordinate (gxE) at ($(rect.west)!0.81!(rect.east)$);
        
          \coordinate (gy1) at ($(rect.south)!0.75!(rect.north)$);
          \coordinate (p1)  at ($(gxW|-gy1)$);
          \fill[#2!60]
            ($(p1)+(-\dia,0)$) -- ($(p1)+(0,\dia)$) -- ($(p1)+(\dia,0)$) -- ($(p1)+(0,-\dia)$) -- cycle;
          \draw[#2!60, line width=\BORDER, line cap=round]
            ($(p1)+(\dia+0.6pt,0)$) -- ([xshift=-2pt]gxE|-gy1);
        
          \coordinate (gy2) at ($(rect.south)!0.50!(rect.north)$);
          \coordinate (p2)  at ($(gxW|-gy2)$);
          \fill[#2!60]
            ($(p2)+(-\dia,0)$) -- ($(p2)+(0,\dia)$) -- ($(p2)+(\dia,0)$) -- ($(p2)+(0,-\dia)$) -- cycle;
          \draw[#2!60, line width=\BORDER, line cap=round]
            ($(p2)+(\dia+0.6pt,0)$) -- ([xshift=-5pt]gxE|-gy2);
        
          \coordinate (gy3) at ($(rect.south)!0.25!(rect.north)$);
          \coordinate (p3)  at ($(gxW|-gy3)$);
          \fill[#2!60]
            ($(p3)+(-\dia,0)$) -- ($(p3)+(0,\dia)$) -- ($(p3)+(\dia,0)$) -- ($(p3)+(0,-\dia)$) -- cycle;
          \draw[#2!60, line width=\BORDER, line cap=round]
            ($(p3)+(\dia+0.6pt,0)$) -- (gxE|-gy3);
        \end{scope}

        \begin{scope}
          \clip (summary.south west) rectangle (summary.north east);
        
          \path coordinate (gxW) at ($(summary.west)!0.18!(summary.east)$);
          \path coordinate (gxE) at ($(summary.west)!0.75!(summary.east)$);
        
          \path coordinate (gy1) at ($(summary.south)!0.75!(summary.north)$);
          \fill[#3!60] (gxW|-gy1) circle[radius=1.2pt];
          \draw[#3!60, line width=\BORDER, line cap=round]
               ($(gxW|-gy1)+(1.6pt,0)$) -- ([xshift=-0pt]gxE|-gy1);
        
          \path coordinate (gy2) at ($(summary.south)!0.50!(summary.north)$);
          \fill[#3!60] (gxW|-gy2) circle[radius=1.2pt];
          \draw[#3!60, line width=\BORDER, line cap=round]
               ($(gxW|-gy2)+(1.6pt,0)$) -- ([xshift=-3pt]gxE|-gy2);
        
          \path coordinate (gy3) at ($(summary.south)!0.25!(summary.north)$);
          \fill[#3!60] (gxW|-gy3) circle[radius=1.2pt];
          \draw[#3!60, line width=\BORDER, line cap=round]
               ($(gxW|-gy3)+(1.6pt,0)$) -- ([xshift=-6pt]gxE|-gy3);
        \end{scope}
        \coordinate (-sqwest)   at (sq.west);
        \coordinate (-rectwest) at (rect.west);
        \coordinate (-recteast) at (rect.east);
        \coordinate (-east)     at (summary.east);
        \coordinate (-north) at (summary.north);
        \coordinate (-rectnorth) at (rect.north);
        \coordinate (-rectsouth) at (rectprmpt.south);
      \end{scope}
    }
  },
  pics/inftythink_iter_n-1/.default={mila-purple-1}{mila-purple-2}
}

\tikzset{
  pics/inftythink_iter_n/.style n args={4}{
    code={
      \begin{scope}[node distance=0pt]
        \begin{scope}[local bounding box=sq]
          \path[draw=none, preaction={fill=#1!10}, pattern color=black!30]
            (\rectr,0) -- (\sww,0) -- (\sww,\sz) -- (\rectr,\sz)
            arc (90:180:\rectr) -- (0,\rectr) arc (180:270:\rectr) -- cycle;
          \draw[#1, dashed, line width=1.4\BORDER]
            (\sww,\sz) -- (\rectr,\sz) arc (90:180:\rectr) -- (0,\rectr)
            arc (180:270:\rectr) -- (\rectr,0) -- (\sww,0);
          \path[use as bounding box] (0,0) rectangle (\sww,\sz);
        \end{scope}
        \node[font=\bfseries\large] at (sq.center) {$\mathrm{q}$};

        \node[
          draw=none, minimum height=\sz, minimum width=0.4\rectw,
          inner sep=0pt, outer sep=0pt, right=0pt of sq
        ] (rectprmpt) {};
        \begin{scope}[on background layer]
          \fill[#3!10] (rectprmpt.south west) rectangle (rectprmpt.north east);
        \end{scope}
        \draw[#3, line width=1.4\BORDER, dashed]
          (rectprmpt.north west) -- (rectprmpt.north east)
          (rectprmpt.north east) -- (rectprmpt.south east)
          (rectprmpt.south east) -- (rectprmpt.south west);

        \begin{scope}
          \clip (rectprmpt.south west) rectangle (rectprmpt.north east);
        
          \path coordinate (gxW) at ($(rectprmpt.west)!0.19!(rectprmpt.east)$);
          \path coordinate (gxE) at ($(rectprmpt.west)!0.81!(rectprmpt.east)$);
        
          \path coordinate (gy1) at ($(rectprmpt.south)!0.75!(rectprmpt.north)$);
          \fill[#3!60] (gxW|-gy1) circle[radius=1.2pt];
          \draw[#3!60, line width=\BORDER, line cap=round]
               ($(gxW|-gy1)+(1.6pt,0)$) -- ([xshift=-0pt]gxE|-gy1);
        
          \path coordinate (gy2) at ($(rectprmpt.south)!0.50!(rectprmpt.north)$);
          \fill[#3!60] (gxW|-gy2) circle[radius=1.2pt];
          \draw[#3!60, line width=\BORDER, line cap=round]
               ($(gxW|-gy2)+(1.6pt,0)$) -- ([xshift=-3pt]gxE|-gy2);
        
          \path coordinate (gy3) at ($(rectprmpt.south)!0.25!(rectprmpt.north)$);
          \fill[#3!60] (gxW|-gy3) circle[radius=1.2pt];
          \draw[#3!60, line width=\BORDER, line cap=round]
               ($(gxW|-gy3)+(1.6pt,0)$) -- ([xshift=-6pt]gxE|-gy3);
        \end{scope}

        \begin{scope}[local bounding box=rect, xshift=3pt+\sw+0.4\rectw]
          \path[
            draw=#2, line width=1.4\BORDER,
            preaction={fill=#2!10}, pattern color=black!30
          ]
          (0.8\rectw,0) -- (0,0) -- (0,\sz) -- (0.8\rectw,\sz);
        \end{scope}

        \begin{scope}[xshift=3pt+\sw+1.2\rectw, local bounding box=conclusion]
          \path[
            draw=#4, line width=1.4\BORDER,
            preaction={fill=#4!10}, pattern color=black!30
          ]
          (0,\sz) -- (\rectwconclu,\sz) 
          arc (90:0:\rectr) -- (\rectwconclu+\rectr,\rectr) arc (0:-90:\rectr) -- (\rectwconclu,0) -- (0,0);
        \end{scope}
        
        \begin{scope}[on background layer]
          \fill[#2!10] (rect.south west) rectangle (rect.north east);
        \end{scope}
        
        \begin{scope}
          \clip (rect.south west) rectangle (rect.north east);
        
          \def\dia{1.8pt}
        
          \coordinate (gxW) at ($(rect.west)!0.19!(rect.east)$);
          \coordinate (gxE) at ($(rect.west)!0.81!(rect.east)$);
        
          \coordinate (gy1) at ($(rect.south)!0.75!(rect.north)$);
          \coordinate (p1)  at ($(gxW|-gy1)$);
          \fill[#2!60]
            ($(p1)+(-\dia,0)$) -- ($(p1)+(0,\dia)$) -- ($(p1)+(\dia,0)$) -- ($(p1)+(0,-\dia)$) -- cycle;
          \draw[#2!60, line width=\BORDER, line cap=round]
            ($(p1)+(\dia+0.6pt,0)$) -- ([xshift=-2pt]gxE|-gy1);
        
          \coordinate (gy2) at ($(rect.south)!0.50!(rect.north)$);
          \coordinate (p2)  at ($(gxW|-gy2)$);
          \fill[#2!60]
            ($(p2)+(-\dia,0)$) -- ($(p2)+(0,\dia)$) -- ($(p2)+(\dia,0)$) -- ($(p2)+(0,-\dia)$) -- cycle;
          \draw[#2!60, line width=\BORDER, line cap=round]
            ($(p2)+(\dia+0.6pt,0)$) -- ([xshift=-5pt]gxE|-gy2);
        
          \coordinate (gy3) at ($(rect.south)!0.25!(rect.north)$);
          \coordinate (p3)  at ($(gxW|-gy3)$);
          \fill[#2!60]
            ($(p3)+(-\dia,0)$) -- ($(p3)+(0,\dia)$) -- ($(p3)+(\dia,0)$) -- ($(p3)+(0,-\dia)$) -- cycle;
          \draw[#2!60, line width=\BORDER, line cap=round]
            ($(p3)+(\dia+0.6pt,0)$) -- (gxE|-gy3);
        \end{scope}

        \begin{scope}
          \clip (conclusion.south west) rectangle (conclusion.north east);
        
          \def\triH{1.2pt}   %
          \def\triL{1.8pt}   %
        
          \coordinate (gxW) at ($(conclusion.west)!0.18!(conclusion.east)$);
          \coordinate (gxE) at ($(conclusion.west)!0.73!(conclusion.east)$);
        
          \coordinate (gy1) at ($(conclusion.south)!0.75!(conclusion.north)$);
          \coordinate (p1)  at ($(gxW|-gy1)$);
          \fill[#4!60]
            ($(p1)+(-\triL,-\triH)$) --
            ($(p1)+(-\triL, \triH)$) --
            ($(p1)+(\triL,0)$) -- cycle;
          \draw[#4!60, line width=\BORDER, line cap=round]
            ($(p1)+(\triL+0.6pt,0)$) -- ([xshift=-3pt]gxE|-gy1);
        
          \coordinate (gy2) at ($(conclusion.south)!0.50!(conclusion.north)$);
          \coordinate (p2)  at ($(gxW|-gy2)$);
          \fill[#4!60]
            ($(p2)+(-\triL,-\triH)$) --
            ($(p2)+(-\triL, \triH)$) --
            ($(p2)+(\triL,0)$) -- cycle;
          \draw[#4!60, line width=\BORDER, line cap=round]
            ($(p2)+(\triL+0.6pt,0)$) -- ([xshift=-5pt]gxE|-gy2);
        
          \coordinate (gy3) at ($(conclusion.south)!0.25!(conclusion.north)$);
          \coordinate (p3)  at ($(gxW|-gy3)$);
          \fill[#4!60]
            ($(p3)+(-\triL,-\triH)$) --
            ($(p3)+(-\triL, \triH)$) --
            ($(p3)+(\triL,0)$) -- cycle;
          \draw[#4!60, line width=\BORDER, line cap=round]
            ($(p3)+(\triL+0.6pt,0)$) -- (gxE|-gy3);
        \end{scope}

        \coordinate (-sqwest)   at (sq.west);
        \coordinate (-rectwest) at (rect.west);
        \coordinate (-recteast) at (rect.east);
        \coordinate (-east)     at (conclusion.east);
        \coordinate (-north) at (conclusion.north);
        \coordinate (-rectnorth) at (rect.north);
        \coordinate (-rectsouth) at (rectprmpt.south);
      \end{scope}
    }
  },
  pics/inftythink_iter_n/.default={mila-purple-1}{mila-purple-2}
}

\tikzset{
  pics/segment/.style n args={2}{
    code={
      \begin{scope}[node distance=0pt]
        \begin{scope}[local bounding box=sq]
          \path[
            draw=delethink-purple, line width=1.4\BORDER, dashed,
            preaction={fill=delethink-purple!5}, pattern color=black!30
          ]
          (\rectr,0) -- (\sw,0) -- (\sw,\sz) -- (\rectr,\sz)
          arc (90:180:\rectr) -- (0,\rectr) arc (180:270:\rectr) -- cycle;
        \end{scope}
        \node[font=\bfseries\large] at (sq.center) {$\mathrm{q}$};

        \node[
          draw=delethink-purple, line width=1.4\BORDER,
          minimum height=\sz, minimum width=\rectw, inner sep=0pt, right=3pt of sq
        ] (rect) {};

        \draw[dashed, line cap=round,
      shorten >=-3pt, shorten <=-3pt](rect.north) -- (rect.south);

        \begin{scope}[on background layer]
          \fill[#1] (rect.south west) rectangle (rect.north);
          \fill[#2] (rect.south) rectangle (rect.north east);
        \end{scope}

        \begin{scope}
          \clip (rect.south west) rectangle (rect.north east);
        
          \def\dia{1.8pt}
        
          \coordinate (gxW) at ($(rect.west)!0.12!(rect.east)$);
          \coordinate (gxE) at ($(rect.west)!0.42!(rect.east)$);
        
          \coordinate (gy1) at ($(rect.south)!0.75!(rect.north)$);
          \coordinate (p1)  at ($(gxW|-gy1)$);
          \fill[delethink-purple!40]
            ($(p1)+(-\dia,0)$) -- ($(p1)+(0,\dia)$) -- ($(p1)+(\dia,0)$) -- ($(p1)+(0,-\dia)$) -- cycle;
          \draw[delethink-purple!40, line width=\BORDER, line cap=round]
            ($(p1)+(\dia+0.6pt,0)$) -- ([xshift=-2pt]gxE|-gy1);
        
          \coordinate (gy2) at ($(rect.south)!0.50!(rect.north)$);
          \coordinate (p2)  at ($(gxW|-gy2)$);
          \fill[delethink-purple!40]
            ($(p2)+(-\dia,0)$) -- ($(p2)+(0,\dia)$) -- ($(p2)+(\dia,0)$) -- ($(p2)+(0,-\dia)$) -- cycle;
          \draw[delethink-purple!40, line width=\BORDER, line cap=round]
            ($(p2)+(\dia+0.6pt,0)$) -- ([xshift=-5pt]gxE|-gy2);
        
          \coordinate (gy3) at ($(rect.south)!0.25!(rect.north)$);
          \coordinate (p3)  at ($(gxW|-gy3)$);
          \fill[delethink-purple!40]
            ($(p3)+(-\dia,0)$) -- ($(p3)+(0,\dia)$) -- ($(p3)+(\dia,0)$) -- ($(p3)+(0,-\dia)$) -- cycle;
          \draw[delethink-purple!40, line width=\BORDER, line cap=round]
            ($(p3)+(\dia+0.6pt,0)$) -- (gxE|-gy3);
        \end{scope}

        \begin{scope}
          \clip (rect.south west) rectangle (rect.north east);
        
          \def\dia{1.8pt}
        
          \coordinate (gxW) at ($(rect.west)!0.585!(rect.east)$);
          \coordinate (gxE) at ($(rect.west)!0.89!(rect.east)$);
        
          \coordinate (gy1) at ($(rect.south)!0.75!(rect.north)$);
          \coordinate (p1)  at ($(gxW|-gy1)$);
          \fill[delethink-purple!40]
            ($(p1)+(-\dia,0)$) -- ($(p1)+(0,\dia)$) -- ($(p1)+(\dia,0)$) -- ($(p1)+(0,-\dia)$) -- cycle;
          \draw[delethink-purple!40, line width=\BORDER, line cap=round]
            ($(p1)+(\dia+0.6pt,0)$) -- ([xshift=-7pt]gxE|-gy1);
        
          \coordinate (gy2) at ($(rect.south)!0.50!(rect.north)$);
          \coordinate (p2)  at ($(gxW|-gy2)$);
          \fill[delethink-purple!40]
            ($(p2)+(-\dia,0)$) -- ($(p2)+(0,\dia)$) -- ($(p2)+(\dia,0)$) -- ($(p2)+(0,-\dia)$) -- cycle;
          \draw[delethink-purple!40, line width=\BORDER, line cap=round]
            ($(p2)+(\dia+0.6pt,0)$) -- ([xshift=-1pt]gxE|-gy2);
        
          \coordinate (gy3) at ($(rect.south)!0.25!(rect.north)$);
          \coordinate (p3)  at ($(gxW|-gy3)$);
          \fill[delethink-purple!40]
            ($(p3)+(-\dia,0)$) -- ($(p3)+(0,\dia)$) -- ($(p3)+(\dia,0)$) -- ($(p3)+(0,-\dia)$) -- cycle;
          \draw[delethink-purple!40, line width=\BORDER, line cap=round]
            ($(p3)+(\dia+0.6pt,0)$) -- ([xshift=-5pt]gxE|-gy3);
        \end{scope}

        \coordinate (-sqwest)   at (sq.west);
        \coordinate (-rectwest) at (rect.west);
        \coordinate (-recteast) at (rect.east);
        \coordinate (-east)     at (rect.east);
        \coordinate (-rectnorth) at ([xshift=0.25\rectw]rect.north);
        \coordinate (-rectsouth) at (rect.south);
      \end{scope}
    }
  },
  pics/segment/.default={mila-purple-1}{mila-purple-2}
}

\tikzset{
  pics/segmentlongcot/.style n args={3}{
    code={
      \begin{scope}[node distance=0pt]
        \begin{scope}[local bounding box=sq]
          \path[
            draw=#1, line width=1.4\BORDER, dashed,
            preaction={fill=#1!10}, pattern color=black!30
          ]
          (\rectr,0) -- (\sw,0) -- (\sw,\sz) -- (\rectr,\sz)
          arc (90:180:\rectr) -- (0,\rectr) arc (180:270:\rectr) -- cycle;
        \end{scope}
        \node[font=\bfseries\large] at (sq.center) {$\mathrm{q}$};

        \begin{scope}[local bounding box=rect, xshift=3pt+\sw]
          \path[
            draw=#2, line width=1.4\BORDER,
            preaction={fill=#2!10}, pattern color=black!30
          ]
          (\rectwlongcot,0) -- (0,0) -- (0,\sz) -- (\rectwlongcot,\sz);
        \end{scope}

        \begin{scope}[xshift=3pt+\sw+\rectwlongcot, local bounding box=conclusion]
          \path[
            draw=#3, line width=1.4\BORDER,
            preaction={fill=#3!10}, pattern color=black!30
          ]
          (0,\sz) -- (\rectwconclu,\sz) 
          arc (90:0:\rectr) -- (\rectwconclu+\rectr,\rectr) arc (0:-90:\rectr) -- (\rectwconclu,0) -- (0,0);
        \end{scope}

        \begin{scope}[on background layer]
          \fill[#2!10] (rect.south west) rectangle (rect.north east);
        \end{scope}

        \begin{scope}
          \clip (rect.south west) rectangle (rect.north east);
        
          \def\dia{1.8pt}
        
          \coordinate (gxW) at ($(rect.west)!0.019!(rect.east)$);
          \coordinate (gxE) at ($(rect.west)!0.065!(rect.east)$);
        
          \coordinate (gy1) at ($(rect.south)!0.75!(rect.north)$);
          \coordinate (p1)  at ($(gxW|-gy1)$);
          \fill[delethink-blue!60]
            ($(p1)+(-\dia,0)$) -- ($(p1)+(0,\dia)$) -- ($(p1)+(\dia,0)$) -- ($(p1)+(0,-\dia)$) -- cycle;
          \draw[delethink-blue!60, line width=\BORDER, line cap=round]
            ($(p1)+(\dia+0.6pt,0)$) -- ([xshift=-2pt]gxE|-gy1);
        
          \coordinate (gy2) at ($(rect.south)!0.50!(rect.north)$);
          \coordinate (p2)  at ($(gxW|-gy2)$);
          \fill[delethink-blue!60]
            ($(p2)+(-\dia,0)$) -- ($(p2)+(0,\dia)$) -- ($(p2)+(\dia,0)$) -- ($(p2)+(0,-\dia)$) -- cycle;
          \draw[delethink-blue!60, line width=\BORDER, line cap=round]
            ($(p2)+(\dia+0.6pt,0)$) -- ([xshift=-5pt]gxE|-gy2);
        
          \coordinate (gy3) at ($(rect.south)!0.25!(rect.north)$);
          \coordinate (p3)  at ($(gxW|-gy3)$);
          \fill[delethink-blue!60]
            ($(p3)+(-\dia,0)$) -- ($(p3)+(0,\dia)$) -- ($(p3)+(\dia,0)$) -- ($(p3)+(0,-\dia)$) -- cycle;
          \draw[delethink-blue!60, line width=\BORDER, line cap=round]
            ($(p3)+(\dia+0.6pt,0)$) -- ([xshift=-0pt]gxE|-gy3);
        \end{scope}

        \begin{scope}
          \clip (rect.south west) rectangle (rect.north east);
        
          \def\dia{1.8pt}
        
          \coordinate (gxW) at ($(rect.west)!0.085!(rect.east)$);
          \coordinate (gxE) at ($(rect.west)!0.130!(rect.east)$);
        
          \coordinate (gy1) at ($(rect.south)!0.75!(rect.north)$);
          \coordinate (p1)  at ($(gxW|-gy1)$);
          \fill[delethink-blue!50]
            ($(p1)+(-\dia,0)$) -- ($(p1)+(0,\dia)$) -- ($(p1)+(\dia,0)$) -- ($(p1)+(0,-\dia)$) -- cycle;
          \draw[delethink-blue!50, line width=\BORDER, line cap=round]
            ($(p1)+(\dia+0.6pt,0)$) -- ([xshift=-6pt]gxE|-gy1);
        
          \coordinate (gy2) at ($(rect.south)!0.50!(rect.north)$);
          \coordinate (p2)  at ($(gxW|-gy2)$);
          \fill[delethink-blue!50]
            ($(p2)+(-\dia,0)$) -- ($(p2)+(0,\dia)$) -- ($(p2)+(\dia,0)$) -- ($(p2)+(0,-\dia)$) -- cycle;
          \draw[delethink-blue!50, line width=\BORDER, line cap=round]
            ($(p2)+(\dia+0.6pt,0)$) -- ([xshift=-0pt]gxE|-gy2);
        
          \coordinate (gy3) at ($(rect.south)!0.25!(rect.north)$);
          \coordinate (p3)  at ($(gxW|-gy3)$);
          \fill[delethink-blue!50]
            ($(p3)+(-\dia,0)$) -- ($(p3)+(0,\dia)$) -- ($(p3)+(\dia,0)$) -- ($(p3)+(0,-\dia)$) -- cycle;
          \draw[delethink-blue!50, line width=\BORDER, line cap=round]
            ($(p3)+(\dia+0.6pt,0)$) -- ([xshift=-4pt]gxE|-gy3);
        \end{scope}

        \begin{scope}
          \clip (rect.south west) rectangle (rect.north east);
        
          \def\dia{1.8pt}
        
          \coordinate (gxW) at ($(rect.west)!0.15!(rect.east)$);
          \coordinate (gxE) at ($(rect.west)!0.19!(rect.east)$);
        
          \coordinate (gy1) at ($(rect.south)!0.75!(rect.north)$);
          \coordinate (p1)  at ($(gxW|-gy1)$);
          \fill[delethink-blue!40]
            ($(p1)+(-\dia,0)$) -- ($(p1)+(0,\dia)$) -- ($(p1)+(\dia,0)$) -- ($(p1)+(0,-\dia)$) -- cycle;
          \draw[delethink-blue!40, line width=\BORDER, line cap=round]
            ($(p1)+(\dia+0.6pt,0)$) -- ([xshift=-0pt]gxE|-gy1);
        
          \coordinate (gy2) at ($(rect.south)!0.50!(rect.north)$);
          \coordinate (p2)  at ($(gxW|-gy2)$);
          \fill[delethink-blue!40]
            ($(p2)+(-\dia,0)$) -- ($(p2)+(0,\dia)$) -- ($(p2)+(\dia,0)$) -- ($(p2)+(0,-\dia)$) -- cycle;
          \draw[delethink-blue!40, line width=\BORDER, line cap=round]
            ($(p2)+(\dia+0.6pt,0)$) -- ([xshift=-9pt]gxE|-gy2);
        
          \coordinate (gy3) at ($(rect.south)!0.25!(rect.north)$);
          \coordinate (p3)  at ($(gxW|-gy3)$);
          \fill[delethink-blue!40]
            ($(p3)+(-\dia,0)$) -- ($(p3)+(0,\dia)$) -- ($(p3)+(\dia,0)$) -- ($(p3)+(0,-\dia)$) -- cycle;
          \draw[delethink-blue!40, line width=\BORDER, line cap=round]
            ($(p3)+(\dia+0.6pt,0)$) -- ([xshift=-6pt]gxE|-gy3);
        \end{scope}

        \begin{scope}
          \clip (rect.south west) rectangle (rect.north east);
        
          \def\dia{1.8pt}
        
          \coordinate (gxW) at ($(rect.west)!0.21!(rect.east)$);
          \coordinate (gxE) at ($(rect.west)!0.26!(rect.east)$);
        
          \coordinate (gy1) at ($(rect.south)!0.75!(rect.north)$);
          \coordinate (p1)  at ($(gxW|-gy1)$);
          \fill[delethink-blue!30]
            ($(p1)+(-\dia,0)$) -- ($(p1)+(0,\dia)$) -- ($(p1)+(\dia,0)$) -- ($(p1)+(0,-\dia)$) -- cycle;
          \draw[delethink-blue!30, line width=\BORDER, line cap=round]
            ($(p1)+(\dia+0.6pt,0)$) -- ([xshift=-5pt]gxE|-gy1);
        
          \coordinate (gy2) at ($(rect.south)!0.50!(rect.north)$);
          \coordinate (p2)  at ($(gxW|-gy2)$);
          \fill[delethink-blue!30]
            ($(p2)+(-\dia,0)$) -- ($(p2)+(0,\dia)$) -- ($(p2)+(\dia,0)$) -- ($(p2)+(0,-\dia)$) -- cycle;
          \draw[delethink-blue!30, line width=\BORDER, line cap=round]
            ($(p2)+(\dia+0.6pt,0)$) -- ([xshift=-6pt]gxE|-gy2);
        
          \coordinate (gy3) at ($(rect.south)!0.25!(rect.north)$);
          \coordinate (p3)  at ($(gxW|-gy3)$);
          \fill[delethink-blue!30]
            ($(p3)+(-\dia,0)$) -- ($(p3)+(0,\dia)$) -- ($(p3)+(\dia,0)$) -- ($(p3)+(0,-\dia)$) -- cycle;
          \draw[delethink-blue!30, line width=\BORDER, line cap=round]
            ($(p3)+(\dia+0.6pt,0)$) -- ([xshift=-0pt]gxE|-gy3);
        \end{scope}

        \begin{scope}
          \clip (conclusion.south west) rectangle (conclusion.north east);
        
          \def\triH{1.2pt}   %
          \def\triL{1.8pt}   %
        
          \coordinate (gxW) at ($(conclusion.west)!0.18!(conclusion.east)$);
          \coordinate (gxE) at ($(conclusion.west)!0.73!(conclusion.east)$);
        
          \coordinate (gy1) at ($(conclusion.south)!0.75!(conclusion.north)$);
          \coordinate (p1)  at ($(gxW|-gy1)$);
          \fill[#3!60]
            ($(p1)+(-\triL,-\triH)$) --
            ($(p1)+(-\triL, \triH)$) --
            ($(p1)+(\triL,0)$) -- cycle;
          \draw[#3!60, line width=\BORDER, line cap=round]
            ($(p1)+(\triL+0.6pt,0)$) -- ([xshift=-3pt]gxE|-gy1);
        
          \coordinate (gy2) at ($(conclusion.south)!0.50!(conclusion.north)$);
          \coordinate (p2)  at ($(gxW|-gy2)$);
          \fill[#3!60]
            ($(p2)+(-\triL,-\triH)$) --
            ($(p2)+(-\triL, \triH)$) --
            ($(p2)+(\triL,0)$) -- cycle;
          \draw[#3!60, line width=\BORDER, line cap=round]
            ($(p2)+(\triL+0.6pt,0)$) -- ([xshift=-5pt]gxE|-gy2);
        
          \coordinate (gy3) at ($(conclusion.south)!0.25!(conclusion.north)$);
          \coordinate (p3)  at ($(gxW|-gy3)$);
          \fill[#3!60]
            ($(p3)+(-\triL,-\triH)$) --
            ($(p3)+(-\triL, \triH)$) --
            ($(p3)+(\triL,0)$) -- cycle;
          \draw[#3!60, line width=\BORDER, line cap=round]
            ($(p3)+(\triL+0.6pt,0)$) -- (gxE|-gy3);
        \end{scope}

        \coordinate (-sqwest)   at (sq.west);
        \coordinate (-rectwest) at (rect.west);
        \coordinate (-recteast) at (rect.east);
        \coordinate (-east)     at (conclusion.east);
        \coordinate (-rectnorth) at ([xshift=0.25\rectw]rect.north);
        \coordinate (-rectsouth) at (rect.south);
      \end{scope}
    }
  },
  pics/segmentlongcot/.default={mila-purple-1}{mila-purple}
}

\tikzset{
  pics/segmentintermediate/.style n args={3}{
    code={
      \begin{scope}[node distance=0pt]
        \begin{scope}[local bounding box=sq]
          \path[draw=none, preaction={fill=delethink-purple!5}, pattern color=black!30]
            (\rectr,0) -- (\sww,0) -- (\sww,\sz) -- (\rectr,\sz)
            arc (90:180:\rectr) -- (0,\rectr) arc (180:270:\rectr) -- cycle;
          \draw[delethink-purple, dashed, line width=1.4\BORDER]
            (\sww,\sz) -- (\rectr,\sz) arc (90:180:\rectr) -- (0,\rectr)
            arc (180:270:\rectr) -- (\rectr,0) -- (\sww,0);
          \path[use as bounding box] (0,0) rectangle (\sww,\sz);
        \end{scope}
        \node[font=\bfseries\large] at (sq.center) {$\mathrm{q}$};

        \node[
          draw=none, minimum height=\sz, minimum width=0.42\rectw,
          inner sep=0pt, outer sep=0pt, right=0pt of sq
        ] (rectprmpt) {};
        \begin{scope}[on background layer]
          \fill[#1] (rectprmpt.south west) rectangle (rectprmpt.north east);
        \end{scope}
        \draw[delethink-purple, line width=1.4\BORDER, dashed]
          (rectprmpt.north west) -- (rectprmpt.north east)
          (rectprmpt.north east) -- (rectprmpt.south east)
          (rectprmpt.south east) -- (rectprmpt.south west);

        \node[
          draw=#3, line width=1.4\BORDER,
          minimum height=\sz, minimum width=0.5\rectw,
          inner sep=0pt, outer sep=0pt, right=4pt of rectprmpt
        ] (rect) {};
        \begin{scope}[on background layer]
          \fill[#2] (rect.south west) rectangle (rect.north east);
        \end{scope}

        \begin{scope}
          \clip (rect.south west) rectangle (rect.north east);
        
          \path coordinate (gxW) at ($(rect.west)!0.18!(rect.east)$);
          \path coordinate (gxE) at ($(rect.west)!0.80!(rect.east)$);
        
          \path coordinate (gy1) at ($(rect.south)!0.75!(rect.north)$);
          \fill[white!60] (gxW|-gy1) circle[radius=1.2pt];
          \draw[white!60, line width=\BORDER, line cap=round]
               ($(gxW|-gy1)+(1.6pt,0)$) -- (gxE|-gy1);
        
          \path coordinate (gy2) at ($(rect.south)!0.50!(rect.north)$);
          \fill[white!60] (gxW|-gy2) circle[radius=1.2pt];
          \draw[white!60, line width=\BORDER, line cap=round]
               ($(gxW|-gy2)+(1.6pt,0)$) -- (gxE|-gy2);
        
          \path coordinate (gy3) at ($(rect.south)!0.25!(rect.north)$);
          \fill[white!60] (gxW|-gy3) circle[radius=1.2pt];
          \draw[white!60, line width=\BORDER, line cap=round]
               ($(gxW|-gy3)+(1.6pt,0)$) -- (gxE|-gy3);
        \end{scope}

        \coordinate (-sqwest)   at (sq.west);
        \coordinate (-rectwest) at (rect.west);
        \coordinate (-recteast) at (rect.east);
        \coordinate (-east)     at (rect.east);
        \coordinate (-rectnorth) at (rect.north);
        \coordinate (-rectsouth) at (rectprmpt.south);
      \end{scope}
    }
  },
  pics/segmentintermediate/.default={mila-purple-1}{mila-purple-2}{mila-purple}
}

\tikzset{
  pics/segmentintermediatecircle/.style n args={3}{
    code={
      \begin{scope}[node distance=0pt]
        \begin{scope}[local bounding box=sq]
          \path[draw=none, preaction={fill=delethink-purple!5}, pattern color=black!30]
            (\rectr,0) -- (\sww,0) -- (\sww,\sz) -- (\rectr,\sz)
            arc (90:180:\rectr) -- (0,\rectr) arc (180:270:\rectr) -- cycle;
          \draw[delethink-purple, dashed, line width=1.4\BORDER]
            (\sww,\sz) -- (\rectr,\sz) arc (90:180:\rectr) -- (0,\rectr)
            arc (180:270:\rectr) -- (\rectr,0) -- (\sww,0);
          \path[use as bounding box] (0,0) rectangle (\sww,\sz);
        \end{scope}
        \node[font=\bfseries\large] at (sq.center) {$\mathrm{q}$};

        \node[
          draw=none, minimum height=\sz, minimum width=0.42\rectw,
          inner sep=0pt, outer sep=0pt, right=0pt of sq
        ] (rectprmpt) {};
        \begin{scope}[on background layer]
          \fill[#1] (rectprmpt.south west) rectangle (rectprmpt.north east);
        \end{scope}
        \draw[delethink-purple, line width=1.4\BORDER, dashed]
          (rectprmpt.north west) -- (rectprmpt.north east)
          (rectprmpt.north east) -- (rectprmpt.south east)
          (rectprmpt.south east) -- (rectprmpt.south west);

        \begin{scope}
          \clip (rectprmpt.south west) rectangle (rectprmpt.north east);
        
          \def\dia{1.8pt}
        
          \coordinate (gxW) at ($(rectprmpt.west)!0.19!(rectprmpt.east)$);
          \coordinate (gxE) at ($(rectprmpt.west)!0.82!(rectprmpt.east)$);
        
          \coordinate (gy1) at ($(rectprmpt.south)!0.75!(rectprmpt.north)$);
          \coordinate (p1)  at ($(gxW|-gy1)$);
          \fill[delethink-purple!40]
            ($(p1)+(-\dia,0)$) -- ($(p1)+(0,\dia)$) -- ($(p1)+(\dia,0)$) -- ($(p1)+(0,-\dia)$) -- cycle;
          \draw[delethink-purple!40, line width=\BORDER, line cap=round]
            ($(p1)+(\dia+0.6pt,0)$) -- ([xshift=-6pt]gxE|-gy1);
        
          \coordinate (gy2) at ($(rectprmpt.south)!0.50!(rectprmpt.north)$);
          \coordinate (p2)  at ($(gxW|-gy2)$);
          \fill[delethink-purple!40]
            ($(p2)+(-\dia,0)$) -- ($(p2)+(0,\dia)$) -- ($(p2)+(\dia,0)$) -- ($(p2)+(0,-\dia)$) -- cycle;
          \draw[delethink-purple!40, line width=\BORDER, line cap=round]
            ($(p2)+(\dia+0.6pt,0)$) -- ([xshift=-0pt]gxE|-gy2);
        
          \coordinate (gy3) at ($(rectprmpt.south)!0.25!(rectprmpt.north)$);
          \coordinate (p3)  at ($(gxW|-gy3)$);
          \fill[delethink-purple!40]
            ($(p3)+(-\dia,0)$) -- ($(p3)+(0,\dia)$) -- ($(p3)+(\dia,0)$) -- ($(p3)+(0,-\dia)$) -- cycle;
          \draw[delethink-purple!40, line width=\BORDER, line cap=round]
            ($(p3)+(\dia+0.6pt,0)$) -- ([xshift=-4pt]gxE|-gy3);
        \end{scope}

        \node[
          draw=#3, line width=1.4\BORDER,
          minimum height=\sz, minimum width=0.5\rectw,
          inner sep=0pt, outer sep=0pt, right=4pt of rectprmpt
        ] (rect) {};
        \begin{scope}[on background layer]
          \fill[#2] (rect.south west) rectangle (rect.north east);
        \end{scope}

        \begin{scope}
          \clip (rect.south west) rectangle (rect.north east);
        
          \path coordinate (gxW) at ($(rect.west)!0.18!(rect.east)$);
          \path coordinate (gxE) at ($(rect.west)!0.75!(rect.east)$);
        
          \path coordinate (gy1) at ($(rect.south)!0.75!(rect.north)$);
          \fill[white!60] (gxW|-gy1) circle[radius=1.2pt];
          \draw[white!60, line width=\BORDER, line cap=round]
               ($(gxW|-gy1)+(1.6pt,0)$) -- ([xshift=-0pt]gxE|-gy1);
        
          \path coordinate (gy2) at ($(rect.south)!0.50!(rect.north)$);
          \fill[white!60] (gxW|-gy2) circle[radius=1.2pt];
          \draw[white!60, line width=\BORDER, line cap=round]
               ($(gxW|-gy2)+(1.6pt,0)$) -- ([xshift=-3pt]gxE|-gy2);
        
          \path coordinate (gy3) at ($(rect.south)!0.25!(rect.north)$);
          \fill[white!60] (gxW|-gy3) circle[radius=1.2pt];
          \draw[white!60, line width=\BORDER, line cap=round]
               ($(gxW|-gy3)+(1.6pt,0)$) -- ([xshift=-6pt]gxE|-gy3);
        \end{scope}

        \coordinate (-sqwest)   at (sq.west);
        \coordinate (-rectwest) at (rect.west);
        \coordinate (-recteast) at (rect.east);
        \coordinate (-east)     at (rect.east);
        \coordinate (-rectnorth) at (rect.north);
        \coordinate (-rectsouth) at (rectprmpt.south);
      \end{scope}
    }
  },
  pics/segmentintermediatecircle/.default={mila-purple-1}{mila-purple-2}{mila-purple}
}

\tikzset{
  pics/segmentintermediatesquare/.style n args={3}{
    code={
      \begin{scope}[node distance=0pt]
        \begin{scope}[local bounding box=sq]
          \path[draw=none, preaction={fill=delethink-purple!5}, pattern color=black!30]
            (\rectr,0) -- (\sww,0) -- (\sww,\sz) -- (\rectr,\sz)
            arc (90:180:\rectr) -- (0,\rectr) arc (180:270:\rectr) -- cycle;
          \draw[delethink-purple, dashed, line width=1.4\BORDER]
            (\sww,\sz) -- (\rectr,\sz) arc (90:180:\rectr) -- (0,\rectr)
            arc (180:270:\rectr) -- (\rectr,0) -- (\sww,0);
          \path[use as bounding box] (0,0) rectangle (\sww,\sz);
        \end{scope}
        \node[font=\bfseries\large] at (sq.center) {$\mathrm{q}$};

        \node[
          draw=none, minimum height=\sz, minimum width=0.42\rectw,
          inner sep=0pt, outer sep=0pt, right=0pt of sq
        ] (rectprmpt) {};
        \begin{scope}[on background layer]
          \fill[#1] (rectprmpt.south west) rectangle (rectprmpt.north east);
        \end{scope}
        \draw[delethink-purple, line width=1.4\BORDER, dashed]
          (rectprmpt.north west) -- (rectprmpt.north east)
          (rectprmpt.north east) -- (rectprmpt.south east)
          (rectprmpt.south east) -- (rectprmpt.south west);

        \begin{scope}
          \clip (rectprmpt.south west) rectangle (rectprmpt.north east);
        
          \def\triH{1.2pt}   %
          \def\triL{1.8
          pt}   %
        
          \coordinate (gxW) at ($(rectprmpt.west)!0.19!(rectprmpt.east)$);
          \coordinate (gxE) at ($(rectprmpt.west)!0.82!(rectprmpt.east)$);
        
          \coordinate (gy1) at ($(rectprmpt.south)!0.75!(rectprmpt.north)$);
          \coordinate (p1)  at ($(gxW|-gy1)$);
          \fill[white]
            ($(p1)+(-\triL,-\triH)$) --
            ($(p1)+(-\triL, \triH)$) --
            ($(p1)+(\triL,0)$) -- cycle;
          \draw[white, line width=\BORDER, line cap=round]
            ($(p1)+(\triL+0.6pt,0)$) -- ([xshift=-3pt]gxE|-gy1);
        
          \coordinate (gy2) at ($(rectprmpt.south)!0.50!(rectprmpt.north)$);
          \coordinate (p2)  at ($(gxW|-gy2)$);
          \fill[white]
            ($(p2)+(-\triL,-\triH)$) --
            ($(p2)+(-\triL, \triH)$) --
            ($(p2)+(\triL,0)$) -- cycle;
          \draw[white, line width=\BORDER, line cap=round]
            ($(p2)+(\triL+0.6pt,0)$) -- ([xshift=-5pt]gxE|-gy2);
        
          \coordinate (gy3) at ($(rectprmpt.south)!0.25!(rectprmpt.north)$);
          \coordinate (p3)  at ($(gxW|-gy3)$);
          \fill[white]
            ($(p3)+(-\triL,-\triH)$) --
            ($(p3)+(-\triL, \triH)$) --
            ($(p3)+(\triL,0)$) -- cycle;
          \draw[white, line width=\BORDER, line cap=round]
            ($(p3)+(\triL+0.6pt,0)$) -- (gxE|-gy3);
        \end{scope}

        \node[
          draw=#3, line width=1.4\BORDER,
          minimum height=\sz, minimum width=0.5\rectw,
          inner sep=0pt, outer sep=0pt, right=4pt of rectprmpt
        ] (rect) {};
        \begin{scope}[on background layer]
          \fill[#2] (rect.south west) rectangle (rect.north east);
        \end{scope}

        \begin{scope}
          \clip (rect.south west) rectangle (rect.north east);
        
          \def\sq{2.2pt} %
          \path coordinate (gxW) at ($(rect.west)!0.19!(rect.east)$);
          \path coordinate (gxE) at ($(rect.west)!0.75!(rect.east)$);
        
          \path coordinate (gy1) at ($(rect.south)!0.75!(rect.north)$);
          \fill[white] ($(gxW|-gy1)+(-.5*\sq,-.5*\sq)$) rectangle ++(\sq,\sq);
          \draw[white, line width=\BORDER, line cap=round]
               ($(gxW|-gy1)+(.5*\sq+0.6pt,0)$) -- ([xshift=-8pt]gxE|-gy1);
        
          \path coordinate (gy2) at ($(rect.south)!0.50!(rect.north)$);
          \fill[white] ($(gxW|-gy2)+(-.5*\sq,-.5*\sq)$) rectangle ++(\sq,\sq);
          \draw[white, line width=\BORDER, line cap=round]
               ($(gxW|-gy2)+(.5*\sq+0.6pt,0)$) -- ([xshift=-5pt]gxE|-gy2);
        
          \path coordinate (gy3) at ($(rect.south)!0.25!(rect.north)$);
          \fill[white] ($(gxW|-gy3)+(-.5*\sq,-.5*\sq)$) rectangle ++(\sq,\sq);
          \draw[white, line width=\BORDER, line cap=round]
               ($(gxW|-gy3)+(.5*\sq+0.6pt,0)$) -- (gxE|-gy3);
        \end{scope}

        \coordinate (-sqwest)   at (sq.west);
        \coordinate (-rectwest) at (rect.west);
        \coordinate (-recteast) at (rect.east);
        \coordinate (-east)     at (rect.east);
        \coordinate (-rectnorth) at (rect.north);
        \coordinate (-rectsouth) at (rectprmpt.south);
      \end{scope}
    }
  },
  pics/segmentintermediatesquare/.default={mila-purple-1}{mila-purple-2}{mila-purple}
}

\tikzset{
  pics/segmentintermediatetriangle/.style n args={3}{
    code={
      \begin{scope}[node distance=0pt]
        \begin{scope}[local bounding box=sq]
          \path[draw=none, preaction={fill=delethink-purple!5}, pattern color=black!30]
            (\rectr,0) -- (\sww,0) -- (\sww,\sz) -- (\rectr,\sz)
            arc (90:180:\rectr) -- (0,\rectr) arc (180:270:\rectr) -- cycle;
          \draw[delethink-purple, dashed, line width=1.4\BORDER]
            (\sww,\sz) -- (\rectr,\sz) arc (90:180:\rectr) -- (0,\rectr)
            arc (180:270:\rectr) -- (\rectr,0) -- (\sww,0);
          \path[use as bounding box] (0,0) rectangle (\sww,\sz);
        \end{scope}
        \node[font=\bfseries\large] at (sq.center) {$\mathrm{q}$};

        \node[
          draw=none, minimum height=\sz, minimum width=0.42\rectw,
          inner sep=0pt, outer sep=0pt, right=0pt of sq
        ] (rectprmpt) {};
        \begin{scope}[on background layer]
          \fill[#1] (rectprmpt.south west) rectangle (rectprmpt.north east);
        \end{scope}
        \draw[delethink-purple, line width=1.4\BORDER, dashed]
          (rectprmpt.north west) -- (rectprmpt.north east)
          (rectprmpt.north east) -- (rectprmpt.south east)
          (rectprmpt.south east) -- (rectprmpt.south west);

        \begin{scope}
          \clip (rectprmpt.south west) rectangle (rectprmpt.north east);
        
          \path coordinate (gxW) at ($(rectprmpt.west)!0.19!(rectprmpt.east)$);
          \path coordinate (gxE) at ($(rectprmpt.west)!0.81!(rectprmpt.east)$);
        
          \path coordinate (gy1) at ($(rectprmpt.south)!0.75!(rectprmpt.north)$);
          \fill[white!60] (gxW|-gy1) circle[radius=1.2pt];
          \draw[white!60, line width=\BORDER, line cap=round]
               ($(gxW|-gy1)+(1.6pt,0)$) -- ([xshift=-0pt]gxE|-gy1);
        
          \path coordinate (gy2) at ($(rectprmpt.south)!0.50!(rectprmpt.north)$);
          \fill[white!60] (gxW|-gy2) circle[radius=1.2pt];
          \draw[white!60, line width=\BORDER, line cap=round]
               ($(gxW|-gy2)+(1.6pt,0)$) -- ([xshift=-3pt]gxE|-gy2);
        
          \path coordinate (gy3) at ($(rectprmpt.south)!0.25!(rectprmpt.north)$);
          \fill[white!60] (gxW|-gy3) circle[radius=1.2pt];
          \draw[white!60, line width=\BORDER, line cap=round]
               ($(gxW|-gy3)+(1.6pt,0)$) -- ([xshift=-6pt]gxE|-gy3);
        \end{scope}

        \node[
          draw=#3, line width=1.4\BORDER,
          minimum height=\sz, minimum width=0.5\rectw,
          inner sep=0pt, outer sep=0pt, right=4pt of rectprmpt
        ] (rect) {};
        \begin{scope}[on background layer]
          \fill[#2] (rect.south west) rectangle (rect.north east);
        \end{scope}

        \begin{scope}
          \clip (rect.south west) rectangle (rect.north east);
        
          \def\triH{1.2pt}   %
          \def\triL{1.8
          pt}   %
        
          \coordinate (gxW) at ($(rect.west)!0.18!(rect.east)$);
          \coordinate (gxE) at ($(rect.west)!0.73!(rect.east)$);
        
          \coordinate (gy1) at ($(rect.south)!0.75!(rect.north)$);
          \coordinate (p1)  at ($(gxW|-gy1)$);
          \fill[white]
            ($(p1)+(-\triL,-\triH)$) --
            ($(p1)+(-\triL, \triH)$) --
            ($(p1)+(\triL,0)$) -- cycle;
          \draw[white, line width=\BORDER, line cap=round]
            ($(p1)+(\triL+0.6pt,0)$) -- ([xshift=-3pt]gxE|-gy1);
        
          \coordinate (gy2) at ($(rect.south)!0.50!(rect.north)$);
          \coordinate (p2)  at ($(gxW|-gy2)$);
          \fill[white]
            ($(p2)+(-\triL,-\triH)$) --
            ($(p2)+(-\triL, \triH)$) --
            ($(p2)+(\triL,0)$) -- cycle;
          \draw[white, line width=\BORDER, line cap=round]
            ($(p2)+(\triL+0.6pt,0)$) -- ([xshift=-5pt]gxE|-gy2);
        
          \coordinate (gy3) at ($(rect.south)!0.25!(rect.north)$);
          \coordinate (p3)  at ($(gxW|-gy3)$);
          \fill[white]
            ($(p3)+(-\triL,-\triH)$) --
            ($(p3)+(-\triL, \triH)$) --
            ($(p3)+(\triL,0)$) -- cycle;
          \draw[white, line width=\BORDER, line cap=round]
            ($(p3)+(\triL+0.6pt,0)$) -- (gxE|-gy3);
        \end{scope}

        \coordinate (-sqwest)   at (sq.west);
        \coordinate (-rectwest) at (rect.west);
        \coordinate (-recteast) at (rect.east);
        \coordinate (-east)     at (rect.east);
        \coordinate (-rectnorth) at (rect.north);
        \coordinate (-rectsouth) at (rectprmpt.south);
      \end{scope}
    }
  },
  pics/segmentintermediatetriangle/.default={mila-purple-1}{mila-purple-2}{mila-purple}
}

\tikzset{
  pics/segmentlegend/.style n args={4}{
    code={
      \begin{scope}[node distance=0pt]
        \begin{scope}[local bounding box=sq]
          \path[draw=none, pattern color=black!30]
            (\rectr,0) -- (\sww,0) -- (\sww,0.5\sz) -- (\rectr,0.5\sz)
            arc (90:180:\rectr) -- (0,\rectr) arc (180:270:\rectr) -- cycle;
          \draw[black!70, dashed, line width=0.8\BORDER]
            (\sww,0.5\sz) -- (\rectr,0.5\sz) arc (90:180:\rectr) -- (0,\rectr)
            arc (180:270:\rectr) -- (\rectr,0) -- (\sww,0);
          \path[use as bounding box] (0,0) rectangle (\sww,0.5\sz);
        \end{scope}

        \node[
          draw=none, minimum height=0.5\sz, minimum width=0.04\rectw,
          inner sep=0pt, outer sep=0pt, right=0pt of sq
        ] (rectprmpt) {};
        \draw[black!70, line width=0.8\BORDER, dashed]
          (rectprmpt.north west) -- (rectprmpt.north east)
          (rectprmpt.north east) -- (rectprmpt.south east)
          (rectprmpt.south east) -- (rectprmpt.south west);
        \node[right=1pt of rectprmpt] (txtprompt)
          {\footnotesize \sffamily \textcolor{black!50}{\texttt{Prompt}}};

        \node[
          draw=black!50, line width=1.3\BORDER,
          minimum height=0.5\sz, minimum width=0.25\rectw,
          inner sep=0pt, outer sep=0pt, right=8pt of txtprompt
        ] (rect) {};
        \node[right=1pt of rect] (response)
          {\footnotesize \sffamily \textcolor{black!50}{\texttt{Response}}};

        \node[
          draw=#1, line width=1.3\BORDER, preaction={fill=#1!10},
          minimum height=0.5\sz, minimum width=0.25\rectw,
          inner sep=0pt, outer sep=0pt, right=15pt of response
        ] (query_rect) {};
        \node[right=1pt of query_rect] (query)
          {\footnotesize \sffamily \textcolor{#1}{\texttt{Query}}};

        \node[
          draw=#2, line width=1.3\BORDER, preaction={fill=#2!10},
          minimum height=0.5\sz, minimum width=0.25\rectw,
          inner sep=0pt, outer sep=0pt, right=8pt of query
        ] (reasoning_rect) {};
        \node[right=1pt of reasoning_rect] (reasoning)
          {\footnotesize \sffamily \textcolor{#2}{\texttt{Reasoning}}};

        \node[
          draw=#3, line width=1.3\BORDER, preaction={fill=#3!10},
          minimum height=0.5\sz, minimum width=0.25\rectw,
          inner sep=0pt, outer sep=0pt, right=8pt of reasoning
        ] (summary_rect) {};
        \node[right=1pt of summary_rect] (summary)
          {\footnotesize \sffamily \textcolor{#3}{\texttt{Summary}}};

        \node[
          draw=#4, line width=1.3\BORDER, preaction={fill=#4!10},
          minimum height=0.5\sz, minimum width=0.25\rectw,
          inner sep=0pt, outer sep=0pt, right=8pt of summary
        ] (conclusion_rect) {};
        \node[right=1pt of conclusion_rect] (conclusion)
          {\footnotesize \sffamily \textcolor{#4}{\texttt{Conclusion}}};
      \end{scope}
    }
  },
  pics/segmentlegend/.default={mila-purple-1}{mila-purple-2}{mila-purple}
}

\newcommand{\subfiglongCoT}{
\begin{tikzpicture}
  \path (0,0) pic (seg1) {segmentlongcot={mila-yellow}{delethink-blue}{plot-green}};
  
  \path[fill=black, fill opacity=0, draw opacity=0]
    ([xshift=-0.6cm, yshift=-0.5\sz]seg1-sqwest)
    rectangle ([yshift=0.5\sz]seg1-sqwest);

  \coordinate (mid23) at ($(seg1-east)!0.5!(seg1-sqwest)$);
  \node[anchor=center, fill=none, draw=none, inner xsep=8pt, minimum height=3cm] at (mid23) {};

  \draw[|-|, line width=0.5, draw=black!100, line cap=round]
    ([xshift=1pt, yshift=-25pt]seg1-sqwest) --
    node[midway, below=7pt, align=center, text width=10cm]
      {
       }
    ([xshift=1pt, yshift=-25pt]seg1-east);

  \node[anchor=south west]
    at ([xshift=5pt, yshift=-23pt]current bounding box.north west)
    {\sffamily \large \textbf{Vanilla Reasoning Paradigm}
     \textcolor{black!60}{}};

  \coordinate (NW) at (current bounding box.north west);
  \draw[opacity=0, line width=0pt] (NW) -- ([xshift=16.5cm]NW);
\end{tikzpicture}
}

\newcommand{\subfigdelethink}{
\begin{tikzpicture}
  \path (0,0) pic (seg1) {inftythink_iter_1={mila-yellow}{delethink-blue}{plot-red}};
  
  \path ([yshift=-0.5\sz, xshift=\segap]seg1-east)
    pic (seg2) {inftythink_iter_n-1={mila-yellow}{delethink-blue}{plot-red}};
  \path ([yshift=-0.5\sz, xshift=\segap]seg2-east)
    pic (seg3) {inftythink_iter_n-1={mila-yellow}{delethink-blue}{plot-red}};
  \path ([yshift=-0.5\sz, xshift=\segap]seg3-east)
    pic (seg4) {inftythink_iter_n={mila-yellow}{delethink-blue}{plot-red}{plot-green}};

  \draw[curvedarrow]
    ([yshift=1pt]seg1-north) to[out=40, in=-130, looseness=1.4] ([yshift=-1pt]seg2-rectsouth);
  \draw[curvedarrow]
    ([yshift=1pt]seg2-north) to[out=40, in=-130, looseness=1.4] ([yshift=-1pt]seg3-rectsouth);
  \draw[curvedarrow]
    ([yshift=1pt]seg3-north) to[out=40, in=-130, looseness=1.4] ([yshift=-1pt]seg4-rectsouth);

  \draw[arrowline, shorten >=2pt, shorten <=2pt]
    ([xshift=\segaparrow]seg1-east) -- ([xshift=-0.2\segaparrow]seg2-sqwest);

  \draw[arrowline, shorten >=2pt, shorten <=2pt]
    ([xshift=\segaparrow]seg2-east) -- node[midway, font=\large, fill=white] {\strut$\cdots$}
    ([xshift=-0.2\segaparrow]seg3-sqwest);

  \draw[arrowline, shorten >=2pt, shorten <=2pt]
    ([xshift=\segaparrow]seg3-east) -- ([xshift=-0.2\segaparrow]seg4-sqwest);

  \draw[|-|, line width=0.5, draw=black!100, line cap=round]
    ([xshift=1pt, yshift=-25pt]seg1-sqwest) --
    node[midway, below=7pt, align=center, text width=4cm]
      {\footnotesize \sffamily \textcolor{black!90}{Iter 1\\}}
    ([xshift=1pt, yshift=-25pt]seg1-east);

   \draw[|-|, line width=0.5, draw=black!100, line cap=round]
    ([xshift=1pt, yshift=-25pt]seg2-sqwest) --
    node[midway, below=7pt, align=center, text width=4cm]
      {\footnotesize \sffamily \textcolor{black!90}{Iter 2\\}}
    ([xshift=1pt, yshift=-25pt]seg2-east);

  \draw[|-|, line width=0.5, draw=black!100, line cap=round]
    ([xshift=1pt, yshift=-25pt]seg3-sqwest) --
    node[midway, below=7pt, align=center, text width=4cm]
      {\footnotesize \sffamily \textcolor{black!90}{Iter n-1}}
    ([xshift=1pt, yshift=-25pt]seg3-east);
  
  \draw[|-|, line width=0.5, draw=black!100, line cap=round]
    ([xshift=1pt, yshift=-25pt]seg4-sqwest) --
    node[midway, below=7pt, align=center, text width=4cm]
      {\footnotesize \sffamily \textcolor{black!90}{Iter n}}
    ([xshift=1pt, yshift=-25pt]seg4-east);

  \path ([xshift=-12cm, yshift=-3.3\sz]seg4-sqwest)
    pic (seg5) {segmentlegend={mila-yellow}{delethink-blue}{plot-red}{plot-green}};

  \node[anchor=south west]
    at ([xshift=5pt, yshift=-20pt]current bounding box.north west)
    {\sffamily \large \textbf{InftyThink Reasoning Paradigm}};

  \coordinate (NW) at (current bounding box.north west);
  \draw[opacity=0, line width=0pt] (NW) -- ([xshift=16.5cm]NW);
\end{tikzpicture}
}

\newcommand{\figmethod}{
\begin{figure*}[t]
    \centering
    \begin{minipage}{0.95\textwidth}
      \centering
      \resizebox{\linewidth}{!}{\subfiglongCoT}\par
      \resizebox{\linewidth}{!}{\subfigdelethink}
    \end{minipage}
    \caption{
    \textbf{InftyThink} reasoning paradigm VS. Vanilla reasoning paradigm. \textbf{Upper panel:} The vanilla reasoning paradigm generates a single, continuous long chain-of-thought in one pass. \textbf{Lower panel:} The InftyThink reasoning paradigm decomposes reasoning into multiple iterative rounds, where consecutive iterations are connected via self-generated global summaries.
    }
    \label{fig:inftythink_overview}
\end{figure*}

}